\pdfoutput=1

\documentclass[11pt,table]{article}

\usepackage[preprint]{acl}

\usepackage{times}
\usepackage{latexsym}
\usepackage{hyperref}

\usepackage[T1]{fontenc}

\usepackage[utf8]{inputenc}
\usepackage{enumitem}
\usepackage{microtype}

\usepackage{inconsolata}

\usepackage{graphicx}
\usepackage{supertabular}
\usepackage{makecell}
\usepackage{multirow}
\usepackage{booktabs}

\title{Enhancing LLM Tool Use with High-quality Instruction Data from Knowledge Graph}

\author{
Jingwei Wang\textsuperscript{1},
Zai Zhang\textsuperscript{1,2},
Hao Qian\textsuperscript{1},
Chunjing Gan\textsuperscript{1},
Binbin Hu\textsuperscript{1}, \\
\bf{Ziqi Liu}\textsuperscript{1},
\bf{Zhiqiang Zhang}\textsuperscript{1},
\bf{Jun Zhou}\textsuperscript{1},
\bf{Bin Shi}\textsuperscript{2},
\bf{Bo Dong}\textsuperscript{3}
\\
\textsuperscript{1}Ant Group, China \\
\textsuperscript{2}School of Computer Science and Technology, Xi’an Jiaotong University, China \\
\textsuperscript{3}School of Distance Education, Xi’an Jiaotong University, China\\
\texttt{wangjingwei.wjw@antgroup.com, zhg.zai@gmail.com} \\
}

\begin{document}
\maketitle
\begin{abstract}
Teaching large language models (LLMs) to use tools is crucial for improving their problem-solving abilities and expanding their applications. However, effectively using tools is challenging because it requires a deep understanding of tool functionalities and user intentions. Previous methods relied mainly on LLMs to generate instruction data, but the quality of these data was often insufficient.
In this paper, we propose a new method that uses knowledge graphs to generate high-quality instruction data for LLMs. Knowledge graphs are manually curated datasets rich in semantic information. We begin by extracting various query pathways from a given knowledge graph, which are transformed into a broad spectrum of user queries. We then translate the relationships between entities into actionable tools and parse the pathways of each query into detailed solution steps, thereby creating high-quality instruction data.
Our experiments show that fine-tuning on just a small sample of this synthetic data can significantly improve the tool utilization and overall capabilities of LLMs.
\end{abstract}

\section{Introduction}

The use of tools is a revolutionary hallmark of advanced intelligence in human civilization \cite{qin2024toollearningfoundationmodels}, deeply expanding the limits of our physical capabilities.
Teaching large language models (LLMs) to use real-world tools is imperative to unleash their potential to solve complex problems with accuracy, efficiency, and automation \cite{schick2024toolformer}.

However, proficiently manipulating tools remains a challenging task for LLMs because it requires a thorough understanding of tool functionalities and a deep insight into varying user intentions.
Recently, some studies have found that instruction tuning can significantly enhance the tool-use capabilities of LLMs \cite{tang2023toolalpaca, qin2023toolllm, liu2024toolace}. These methods typically start by collecting real APIs or synthesizing simulated APIs. They then use LLMs to generate user queries based on the APIs. After that, LLMs are employed again to synthesize detailed solution steps for each query, including tool invocations.

Despite the progress made by these methods, several limitations in the construction of instruction data may hinder their full potential, including unguaranteed data quality, insufficient query complexity, and prohibitive costs.
First, data quality is crucial for instruction-tuning-based methods and directly impacts the capabilities of LLMs \cite{gunasekar2023textbooks,zhou2024lima}. Researchers often use advanced commercial models like ChatGPT or GPT-4 to generate data. However, even with these models or careful human review, errors can still occur in the dataset. Second, many methods \cite{qin2023toolllm, srinivasan2023nexusraven} randomly sample APIs and prompt LLMs to generate queries. This simple approach often leads to irrelevant tool combinations and low-complexity queries rather than high-complexity ones that require advanced reasoning skills. As a result, the final dataset may not be challenging enough to fully engage the reasoning and planning capabilities of LLMs. Moreover, the inconsistent quality of LLM-generated data necessitates meticulous manual review and rigorous revision. The extensive human intervention throughout the data construction process—from initial creation to final validation—makes rapid scaling impractical and labor costs prohibitive.

In this paper, we tackle these challenges by using knowledge graphs (KGs) to create high-quality instruction-tuning data. KGs contain rich, structured knowledge with concepts and relationships represented as nodes and edges. From a tool use perspective, the basic unit of KGs—the "entity-relation-entity" triple—can be interpreted as "input-function-output." By extracting subgraphs from KGs, we can generate complex tool combinations that exhibit \textbf{high complexity}, along with corresponding queries and solution paths.
Since KGs are carefully curated and verified by humans, their accuracy and reliability are well-established, ensuring the integrity of the extracted subgraphs. By converting these subgraphs into natural language through a simple formatting process, we can easily generate tool-using instructions and solution paths without relying on potentially flawed LLM-generated data. This approach avoids errors and noise, maintaining the \textbf{high quality} of our datasets.
Moreover, our method bypasses labor-intensive prompting and eliminates redundant interactions with LLMs. Leveraging existing large-scale KGs and using diverse sampling patterns without manual verification, our approach provides an efficient, \textbf{low-cost} solution for scaling up datasets.

In particular, we present a new framework that generates high-quality instruction data by leveraging query-solution pairs from KGs. Our approach integrates First-Order Logic (FOL) queries into the data generation process, ensuring precise execution of each step and guaranteeing answer quality. The framework extracts subgraphs from KGs that match predefined FOL patterns, representing tool utilization queries and solution paths. By executing API sequences associated with these queries, we log API calls and outcomes, creating solution paths and an instruction-tuning dataset called KG2Tool. By fine-tuning various LLMs with KG2Tool, we observe significant performance improvements on the T-Eval benchmark. Our framework thus provides a high-quality, low-cost solution for enhancing the utilization of LLM tools.

Our major contributions are as follows:
\begin{itemize}[leftmargin=*, itemsep=0pt]
    \item We propose to utilize knowledge graphs to generate high-quality instruction data to enhance LLMs' tool use capability.
    \item We design a framework that uses FOL queries as intermediaries to transform KGs into tool-use format APIs, queries, and solution paths.
    \item Extensive experiments with various LLMs validate the effectiveness of our synthesized data.
\end{itemize}

\section{Related Work}
\label{rela_work}
\subsection{Tool Use of LLMs}
Integrating external tools within LLMs has emerged as a growing field of research \citep{qin2024toollearningfoundationmodels}. Current methodologies can be delineated into two discrete lines.
The first line of methods stimulates the tool-use capabilities within LLMs via pure prompting strategies, enabling full interactions among language models, users, and tools. 
These endeavors encompass the utilization of LLMs with a diverse array of tools, such as code interpreters \citep{gao2023pal}, search engines \citep{yao2022react}, retrieval frameworks \citep{khattab2022demonstrate}, etc. \citep{shen2024hugginggpt,lu2024chameleon}.
The remaining methods in the second line enhance tool utilization abilities within LLMs using supervised fine-tuning (SFT).
They commonly leverage closed-sourced LLMs like ChatGPT to construct instruction-tuning datasets tailored for tool usage.

Meanwhile, Retrieval-Augmented Generation (RAG) \cite{lewis2020retrieval} techniques have also been integrated into studying tool learning within LLMs.
While some may opt to employ a retriever to enhance the prompting method \citep{yuan2023craft} directly, other approaches incorporate retrieval components and tool-use tuning procedures via API retrieval \citep{qin2023toolllm} or prompt demonstration \citep{srinivasan2023nexusraven}.

\subsection{Tool-use Instruction Dataset}
Instruction tuning relies heavily on curated datasets to enhance LLMs' capacity to comprehend human instructions and generate appropriate responses \citep{wei2021finetuned,bach2022promptsource}.
Thus, constructing instruction data is the core of tool learning methods based on instruction tuning.
The construction process usually consists of three phases: API collection, query generation, and solution path annotation.
ToolAlpaca \citep{tang2023toolalpaca} simulates an environment to generate tool-use instances without manual intervention. 
ToolFormer \citep{schick2024toolformer} meticulously designs a bootstrapping framework including in-context learning (ICL) prompting, API calls sampling, executing, and filtering to generate an interleaved dataset with API invocations.
Llama3 \citep{dubey2024llama} utilizes a combination of human preference annotations and manual rewrites progressively to generate annotation data.
ToolHop \cite{Ye2025ToolHop} utilizes a question-driven data scheme, employing GPT-4o throughout its three core stages: tool creation, document optimization, and code generation, to facilitate the process.
However, these data construction methods inevitably either involve costly human annotations or heavily rely on unreliable LLM generation, leading to high expenses or unguaranteed quality. Our approach, in contrast, could address these issues.

\begin{figure*}[t!]
\centering
\includegraphics[width=\linewidth]{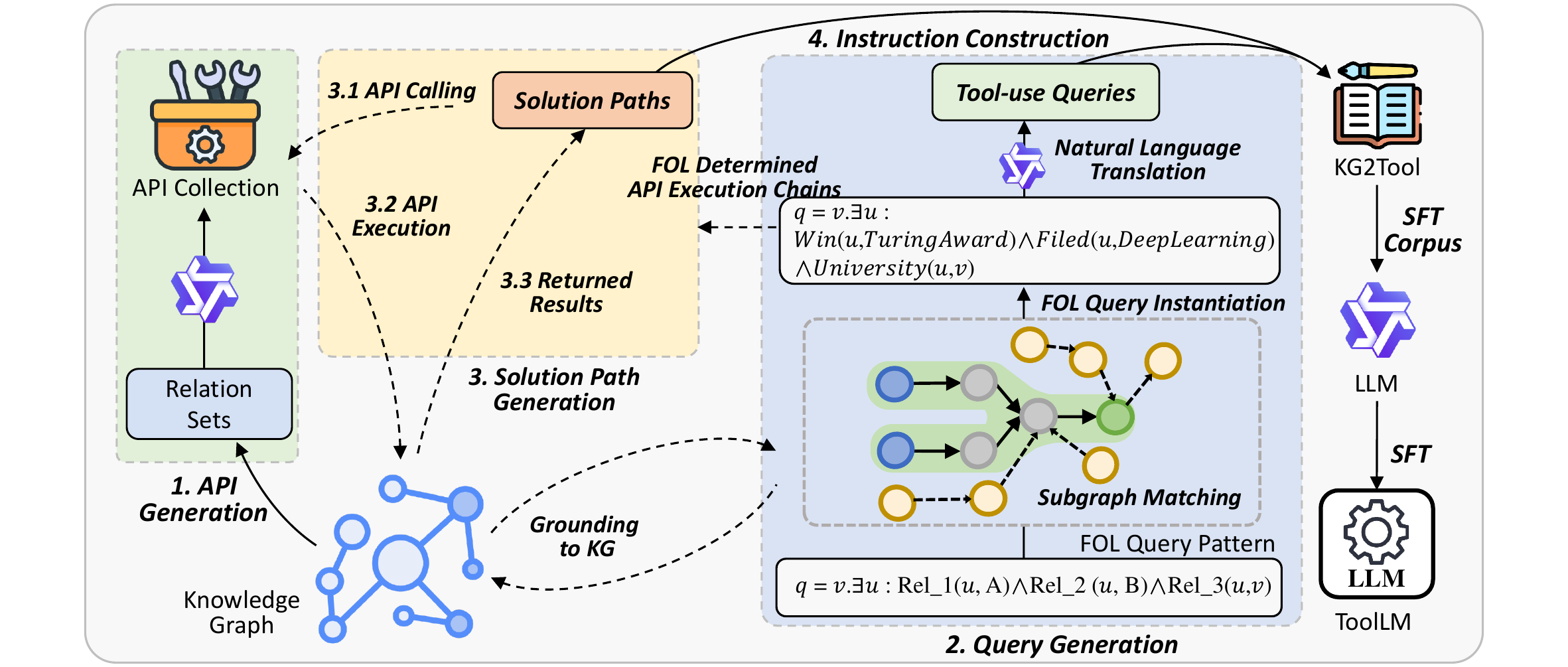}
\caption{Overview of our proposed framework for constructing the KG2Tool dataset. 
The process is divided into three main stages: (1) \textbf{API Generation}, where relation types from the knowledge graph are converted into executable APIs; (2) \textbf{Query Generation}, where FOL query patterns are instantiated via subgraph matching and translated into natural language queries; and (3) \textbf{Solution Path Generation}, where the FOL queries determine API execution chains. These APIs are sequentially called and executed over the knowledge graph, and their returned results are combined to constitute complete and verified solution paths.
}
\label{fig:framework}
\end{figure*}

\subsection{KG for LLMs}

Incorporating KGs can significantly enhance LLMs' ability to acquire up-to-date information and factual knowledge, thereby reducing hallucination. Recent studies have primarily focused on the RAG framework, which retrieves relevant KG subgraphs and integrates them into the input context using well-designed prompts. For example, RoG \cite{luo2023reasoning} proposes a planning-retrieval-reasoning framework to generate relation paths from KGs, enabling valid reasoning for LLMs. StructGPT \cite{jiang2023structgpt} introduces an iterative reading-then-reasoning framework that allows LLMs to access structured knowledge through specialized interfaces, aiding answer generation.
Beyond prompt-based methods, \cite{zhou2024enhancing} shows that SFT with synthetic graph-based reasoning data can effectively improve LLMs' reasoning performance. \cite{wang2024learning} constructs planning data from KGs and fine-tunes LLMs to follow instructions and execute plans to obtain final answers.
To our knowledge, our approach is the first to generate instruction-tuning data from KGs specifically for enhancing LLM tool utilization.

\section{Preliminary}
In this section, we describe the background information of KGs and FOL queries.

\vspace{2mm}

\noindent\textbf{Knowledge Graphs.} 
KGs organize and store real-world knowledge in a heterogeneous graph structure, representing entities as nodes and relations as edges.
Given a set of entities $\mathcal{V}$ and a set of relations $\mathcal{R}$, a knowledge graph can be denoted as a tuple $\mathcal{G} = (\mathcal{V}, \mathcal{E}, \mathcal{R})$, where $\mathcal{E}$ is a set of triplets $\mathcal{E} = {(h_i, r_i, t_i)} \subseteq \mathcal{V} \times \mathcal{R} \times \mathcal{V}$. 
Each triplet represents a fact from head entity $h_i$ to tail entity $t_i$ with the relation type $r_i$.

\vspace{2mm}

\noindent\textbf{First-order Logic Queries.} 
To enable large-scale subgraph sampling from KGs and directly translate these subgraphs into natural language instructions, we introduce FOL queries.
A FOL query is a formula composed of constants (denoted with the entity's name), variables (denoted with lowercase letters, e.g., u, v), relations (denoted with relation term, formatted as R(a, b)) and logic symbols (including $\exists$, $\wedge$, $\vee$, $\neg$).
In our method, each constant or variable represents an entity from the set $\mathcal{V}$.
Every relation symbol R(a, b) acts as a binary function, signifying whether a relation R exists between a pair of constants or variables.
Regarding logical symbols, our considerations encompass conjunction ($\wedge$), disjunction ($\vee$), negation ($\neg$) and existential quantification ($\exists$).
A bounded variable is free if it is quantiﬁed in the expression
If a variable is quantiﬁed with an existential symbol, it is termed a bounded variable; otherwise, it is a free variable.
For example, a natural language question, “Which universities do the Turing Award winners of deep learning work in?” can be equivalent to a FOL query as $q=v.\exists u: Win(u,Turing Award)\wedge Filed(u,Deep Learning) \wedge University(u,v)$ \cite{zhu2022neural}.
For an FOL query, our goal is to find the answers to the free variables that make the formula true.

\vspace{2mm}

\noindent\textbf{Basic Operations on KGs.} 
We define four basic entity operations aligned with relations and logical symbols in FOL queries to enable KG reasoning:
\begin{itemize}[leftmargin=*, itemsep=0pt]
\item 
\textbf{Relation Projection}, denoted as $P_q(A)$, computes the entity set of tail entities reachable by head entities through relation $q$. 
To derive head entities given tail entities, the projection is denoted as $P_{q^{-1}}(A)$ for the inverse relation $q^{-1}$.
\item
\textbf{Intersection Operation}, denoted as $A \cap B$, computes common entities between sets $A$ and $B$. 
\item 
\textbf{Union Operation}, denoted as $A \cup B$, computes the set of entities that belong to either set $A$ or set $B$, or to both.
\item 
\textbf{Complement Operation}, denoted as $U \setminus A$, computes the entities in set $U$ but not in set $A$. $U$ stands for the universal set.
\end{itemize}

\section{Data Construction}
\label{methods}

Drawing on the analogy between relations in triples and function operations, we abstract head nodes and their relations as input parameters and API calls. Building on this, we treat the search for missing nodes in a KG subgraph and its deduction process as the tool-use query and solution path, respectively.
We introduce FOL queries as intermediaries for generating these queries and solution paths. FOL queries naturally align with problem decomposition and multi-step solutions and can be easily translated into natural language. Each FOL query corresponds to a unique subgraph pattern, enabling large-scale sampling of reasoning subgraphs.
By leveraging these advantages, we minimize human effort (such as manual verification and modification) while maximizing data quality and generation efficiency.
As illustrated in Figure \ref{fig:framework}, we will detail our data construction method using FOL queries, covering query generation, API generation, solution path generation, and instruction data construction.

\begin{figure*}[t!]
\centering
\includegraphics[width=\linewidth]{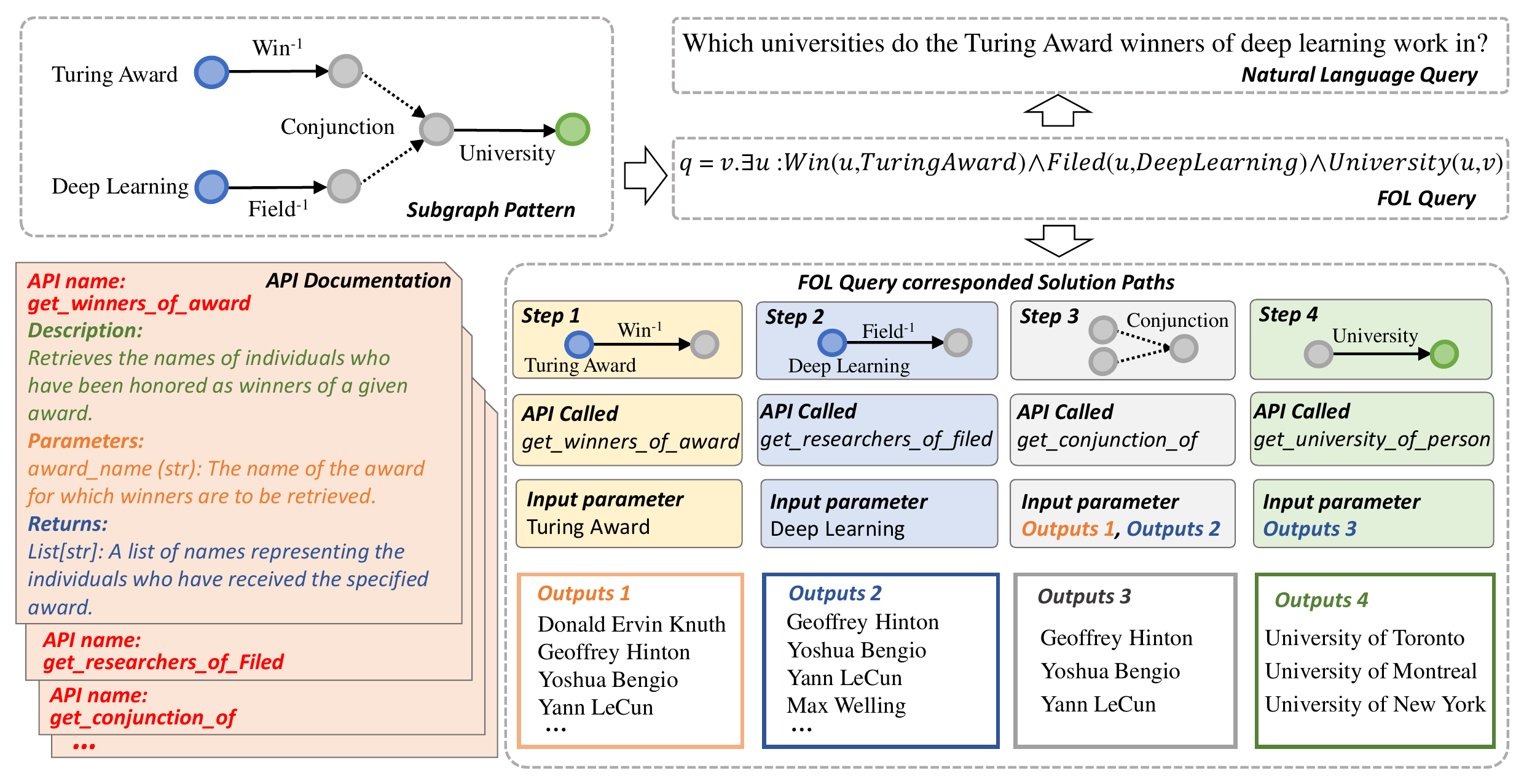}
\caption{Demo of solution path generation from a given subgraph.}
\label{fig:solution_path}
\end{figure*}

\subsection{API Generation}

When written in the projection form $P_{r_i}(h_i)=t_i$, each fact triplet $(h_i, r_i, t_i)$ in the knowledge graph can be understood as applying a relation-specific operation on the head node to yield the tail node.
This closely parallels \textit{function calling} in tool-using, where a function identified by its name is executed on the provided input parameters and returns outputs after performing its designated operations.
Viewed from a translation perspective \cite{bordes2013translating}, each relation instance within a triplet acts as a function that conducts a transformation from its head into its tail.
Therefore, we abstract each relation type as an API in our approach, requiring head nodes as input parameters and returning tail nodes.
While these functions are not directly collected from real-world API repositories,  almost all correspond to genuine needs and functionalities in certain real scenarios.
It should be noted that instead of simulating executions as ToolAlpaca \cite{tang2023toolalpaca}, each API call can get accurate and verifiable results via truly executing queries in the knowledge graph.

To complete the API generation using relation types, we perform format conversions to align with API conventions and linguistic conversions to ensure appropriate function names for projections from head entities to tail entities. 
In most cases, this API generation process is straightforward by simply prefixing \textit{"get\_"} and postfixing \textit{"\_of\_"} with \textit{head type}. 
In the aforementioned query scenario, with "University" as the relation and persons as head entities,  the API name can be easily derived as "get\_university\_of\_person".
When dealing with reverse relations like $Win^{-1}$, it requires generating an API with the inverse meaning of the original relation(i.e., winners).
In practice, we design forward and reverse relation templates, utilizing in-context learning methods to prompt the LLM to generate APIs.
Our API generation approach avoids excessive reasoning that could introduce errors or inaccuracies.

Our API set encompasses relation-based APIs and three dedicated APIs for executing logical operations, including conjunction, disjunction, and negation.
Specifically, \textit{get\_intersection\_of}, \textit{get\_union\_of}, and \textit{get\_negation\_of} serve as the corresponding APIs to perform intersection operation, union operation, and complement operation.
These APIs are implemented as functions and executed by Python interpreters.

\subsection{FOL Instantiation and Query Generation}
To facilitate large-scale data generation of diverse tool-usage queries with reasoning complexity, we use FOL queries as the intermediate form to produce queries and solution paths to ensure efficient and high-quality data generation.
Following prior research \cite{zhu2022neural}, we define a total of 14 FOL query patterns in practice, denoted as $P =
\{1p, 2p, 3p, 2i, 3i, pi, ip, 2u, up, 2in, 3in, inp, pin,$ $ pni\}$.
Here, $p$, $i$, $u$, and $n$ represent operations of relation projection, intersection, union, and complement.
Each  FOL query pattern corresponds to a specific sub-graph pattern, as shown in Appendix Tables \ref{tab:pattern_part1} and \ref{tab:pattern_part2} with their examples.

To instantiate a FOL query pattern using its subgraph pattern, we sample real reasoning subgraphs from the KG via simple subgraph matching. Specifically, we randomly select an entity $e$ from the KG as the root node of the tree-structured subgraph pattern. We then use a pre-order traversal to assign KG entities and relations to the subgraph structure. We uniformly select an existing relation $r$ from $e$'s incoming relations for each edge and assign $r$ and its head entity $e'$ to the corresponding edge and node in the subgraph pattern. This process continues until all nodes and relations are matched. If any required relation is missing, the instantiation fails. We use a post-order traversal for FOL subgraphs with negations to avoid cases where the complement operation results in an empty set. Once successfully instantiated, we fill the entities and relations into the FOL query pattern to obtain the instantiated FOL query. With the help of LLMs, these FOL queries can then be easily transformed into natural language tool-use queries.

Given that tool-use queries are natural language form of our obtained FOL queries, we can easily translate an FOL query (e.g., $q=v.\exists u: Win(u, Turing Award)\wedge Filed(u, Deep Learning) \wedge University(u,v)$) into the final query (e.g., "Which universities do the Turing Award winners of deep learning work in?"), while ensuring the accuracy of this procedure.
We present our prompt for query generation in the Appendix.

\subsection{Solution Path Generation}
Since FOL queries of the same pattern share identical execution sequences, we obtain the execution chains for each FOL pattern by traversing the instantiated subgraph structure in a post-order manner. With the FOL queries established, their solution paths are naturally derived by calling the APIs and recording their results. Because each API call and its response are sourced from reliable knowledge graphs, the correctness of the solution paths is ensured.
For example, consider the instance in Figure \ref{fig:solution_path}. The FOL query dictates the decomposition steps and overall execution sequence. Initially, two APIs are called to retrieve single-step results: one for querying the winners of the \textit{Turing Award} and another for researchers in \textit{Deep learning}. Next, an intersection API is invoked to find overlapping researchers. Finally, a relation projection API is applied to the intermediate set to obtain the desired answers regarding their universities.

\subsection{Instruction Data Construction}
At this point, we can obtain the initial query-solution pairs, where the solution includes detailed steps to resolve the query, with each step containing the required API, parameters, and the API's return results. 
The instruction data for fine-tuning LLMs is typically presented in a chat format. Hence, we convert the initial query-solution pairs into a dialogue format to align with standard instruction data formats (such as Alpaca or ShareGPT). Following \cite{chen2023teval, chen2024agentflan}, we add a system prompt to each query-solution pair, informing the assistant which tools can be called, then regard the query as the user's input and each tool invocation as output, thus constructing complete instruction-following data. Please see Appendix \ref{a3} for more details and examples of the constructed instruction data.
Since we use high-quality KGs like FB15k \cite{kadlec2017knowledge} that have been manually curated, the instruction data constructed in this way does not need to be quality-checked by advanced LLMs like GPT-4, also avoiding the need for extensive manual quality control. We refer to this instruction data constructed from the KG as KG2Tool, and we will make our data publicly available.

\section{Evaluation and Results}
\label{others}

\begin{table*}[h!]
\centering
\caption{Main Results of T-Eval. Overall stands for the score calculated from an average of metrics on all subsets.}
\label{tab:teval}
\begin{tabular}{lccccccc}
\toprule
\textbf{Model}     & \textbf{Instruct} & \textbf{Plan} & \textbf{Reason} & \textbf{Retrieve} & \textbf{Understand} & \textbf{Review} & \textbf{Overall} \\
\midrule
GPT-3.5            & 96.60    & 86.60 & 67.75  & 92.25    & 85.50      & 75.60  & 84.05   \\
GPT-4              & 96.30    & 87.80 & 65.35  & 88.95    & 85.75      & 94.50  & 86.44   \\ 
GPT-4o            & 94.01 & 75.54 & 66.56 & 78.86 & 81.73 & 83.84 & 80.09 \\
\midrule
ToolAlpaca-7B     & 0.47     & 17.26 & 19.07  & 17.73    & 19.52      & 0      & 12.34   \\
LLaMA2-7B          & 34.45    & 28.05 & 22.10  & 16.90    & 24.45      & 38.60  & 27.43   \\
ToolACE-8B        & 2.15     & 37.15 & 26.23  & 17.03    & 17.9       & 68.79  & 28.21   \\
InternLM-7B        & 39.15    & 55.40 & 36.90  & 47.15    & 50.30      & 46.20  & 45.85   \\
ChatGLM3-6B        & 72.05    & 42.70 & 36.15  & 45.25    & 57.75      & 54.80  & 51.45   \\
Mistral-7B         & 61.65    & 71.05 & 39.15  & 51.80    & 48.95      & 63.20  & 55.97   \\
Baichuan2-7B       & 73.00    & 52.30 & 41.30  & 51.10    & 59.65      & 61.40  & 56.46   \\
Qwen-7B            & 61.45    & 64.65 & 45.25  & 62.15    & 61.90      & 61.60  & 59.50   \\
LLaMA3.1-8B        & 81.62    & 69.92 & 66.66  & 76.76    & 73.38      & 82.14  & 75.08   \\
GLM4-9B            & 90.18    & 77.09 & 68.38  & 77.08    & 74.71      & 64.89  & 75.39   \\
Qwen2-7B           & \textbf{97.66}    & 83.02 & 63.92  & 85.65    & 83.41      & 42.51  & 76.03   \\
Qwen2.5-7B         & 93.70    & 74.61 & 66.52  & 87.96    & 75.63      & 67.97  & 77.73   \\
\rowcolor{blue!10} ToolLM-7B  & 97.36    & \textbf{83.83} &  \textbf{75.47}  & \textbf{90.45}    & \textbf{84.64}      & \textbf{76.59}  & \textbf{84.72}   \\
\midrule
LLaMA2-13B         & 33.35    & 56.90 & 26.45  & 24.65    & 29.40      & 53.00  & 37.29   \\
Baichuan2-13B      & 29.85    & 60.80 & 41.85  & 55.70    & 56.00      & 57.30  & 50.25   \\
Qwen-14B           & 73.65    & 74.65 & 52.35  & 75.60    & 64.65      & 56.90  & 66.30   \\
Qwen2.5-14B        & 98.37    & \textbf{87.90} & 69.00  & 84.50    & 77.66      & 74.33  & 81.96   \\
\rowcolor{blue!10} ToolLM-14B & \textbf{98.68}    & 86.68 & \textbf{77.71}  & \textbf{91.96}    & \textbf{85.28}      & \textbf{82.96}  & \textbf{87.21}   \\
\midrule
LLaMA2-70B         & 78.95    & 60.55 & 31.10  & 39.55    & 44.80      & 62.80  & 52.96   \\
Qwen-72B           & 63.05    & 79.25 & 59.45  & 70.90    & 75.30      & 80.30  & 71.38   \\
Qwen2-72B          & 98.36    & 86.29 & 71.06  & 89.95    & 86.87      & 68.17  & 83.45   \\
Qwen2.5-72B        & 98.75    & 88.61 & 72.42  & 90.71    & 80.63      & 89.12  & 86.71   \\
LLaMA3.1-70B       & 98.25    & 89.79 & 69.55  & 91.05    & 88.83      & 83.78  & 86.87   \\
\bottomrule
\end{tabular}
\end{table*}

\subsection{Experimental Setup}

\textbf{Benchmark.} 
We use the largest available tool utilization benchmark to evaluate the tool use performance of LLMs comprehensively: \textbf{T-Eval} \cite{chen2023teval}. T-Eval has 23,305 test cases covering various tool sets and yields 5.8 calling steps for each query on average. T-Eval is a step-by-step tool evaluation benchmark for LLMs, which explicitly decomposes the evaluation into six sub-tasks (i.e., plan, reason, retrieve, understand, instruct, and review) along the basic capabilities of LLMs.

\vspace{2mm}

\noindent
\textbf{Training setting.} 
To test the effectiveness of our data, we conduct extensive experiments by training LLMs with the generated KG2Tool instruction data. We train the open-source LLMs, Qwen2.5-Instruct series \cite{qwen2.5}, in the SFT manner. We refer to the model trained with our data as ToolLM-xB. For example, Qwen2.5-7B, after being fine-tuned with KG2Tool, is denoted as ToolLM-7B.
Due to the limited resources, we adopt the parameter-efficient training strategy LoRA \cite{hu2021lora} to fine-tune all models. As for the hyperparameter setting, we adopt one of the most common settings, which sets the rank as 16 and alpha as 32 for all models. The learning rate is $0.0001$, warmup ratio $0.1$, LR scheduler $cosine$, and batch size $32$. We used 2k randomly sampled data from the KG2Tool for instruction fine-tuning in all experiments.

\vspace{2mm}

\noindent
\textbf{Inference setting.} 
We adopt one of the most popular LLM inference engines, vLLM \cite{kwon2023vllm}, to implement the inference process of various open-source LLMs.
For the inference parameters, we set the temperature to $0$, top\_p to $1.0$, top\_k to $-1$, and batch size to $32$.

\subsection{Overall Performance}
Table \ref{tab:teval} summarizes the results of various LLMs, including closed-source and open-source models, and our models on T-Eval. We divide the results into four groups. The GPT series of proprietary LLMs form one group, while open-source LLMs are divided into three groups based on the scale of their parameters. Within each group, the models are sorted according to their overall performance from lowest to highest. The results of our models are highlighted in light blue. In particular, the best results for each metric in their two groups are highlighted in bold. Below are the key observations derived from the table.

Firstly, closed-source LLMs achieve exceptionally good tool-use performance and have a significant lead compared to early open-source LLMs. However, this gap is rapidly narrowing; for instance, the overall performance of Qwen2-7B has already far surpassed that of previous larger models (like Llama2-70B and Qwen-72B). The overall performance of Qwen2-72B (83.45) and Qwen2.5-72B (86.71) has reached or even surpassed the levels of GPT-3.5 (84.05) and GPT-4 (86.44). These results show that open-source LLMs have made significant progress in tool usage capabilities, and the gap between models with 7B parameters and models with 70B parameters is also narrowing.

Further, we can see that the models fine-tuned with our KG2Tool data have experienced a very significant performance improvement. For instance, ToolLM-7B improved its performance by 9.0\% over Qwen2.5-7B, and its overall score (84.72) surpassed GPT-3.5. In particular, ToolLM-14B achieved the highest score of 87.21, surpassing its 70B counterpart in the same series, Qwen2.5-72B (86.71). These results demonstrate the effectiveness of our KG2Tool data for tool use.

\subsection{Scaling Performance of Model Size}
Generally, the performance of LLMs increases with scale, as shown in Table \ref{tab:teval}. To explore whether models fine-tuned with our data exhibit a similar scale effect, we conduct experiments using the Qwen2.5-Instruct series, which offers a wide range of model sizes. Due to resource limitations, we only performed instruction fine-tuning on models within the scale range of 0.5B to 14B. The original Qwen2.5-Instruct models are denoted as raw models, and the models after instruction fine-tuning with our data are marked as SFT models.

\begin{figure}[t!]
\centering
\includegraphics[width=0.95\linewidth]{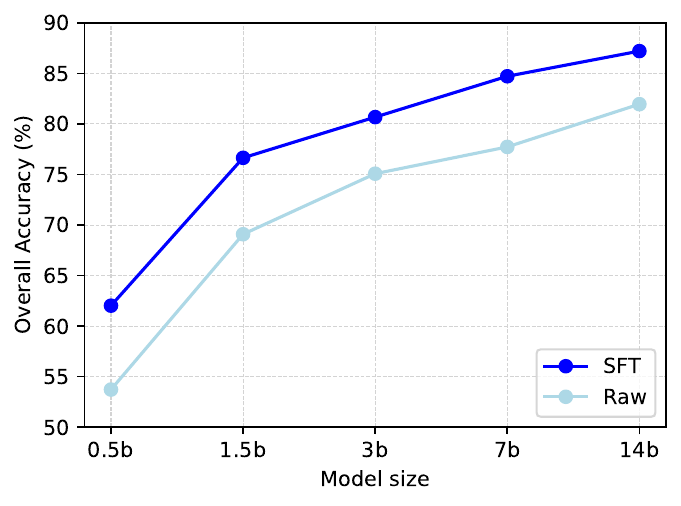}
\caption{Performance scaling laws for the parameters of training models, from 0.5B to 14B.}
\label{fig:model_scaling}
\end{figure}

Both raw and SFT models are evaluated on T-Eval, with results in Figure \ref{fig:model_scaling}. The performance of raw and SFT models continues to improve as the model size increases. At the same time, the advantage of the SFT models over the raw models remains consistent, indicating that our data has the potential to enhance the tool usage capabilities of larger LLMs. Notably, the 3B model for mobile applications scored over 80 after SFT, achieving performance comparable to GPT-4o (80.09). This result suggests that small language models can also achieve tool use performance comparable to advanced LLMs like GPT-4o. This finding strongly supports the development of powerful LLM applications for mobile devices.

\subsection{Study on Various Backbone LLMs}

\begin{figure}[t!]
\centering
\includegraphics[width=0.95\linewidth]{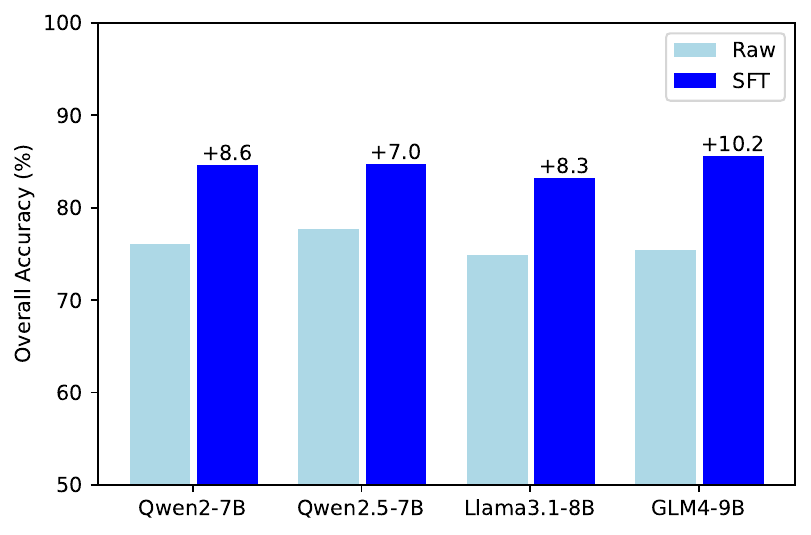}
\caption{Performance of various backbone LLMs fine-tuned with KG2Tool.}
\label{fig:various_llm}
\end{figure}

To verify whether our data is also effective for other LLMs, we selected two other models that are similar in size to Qwen2.5-7B and widely followed by the community (i.e., Llama3.1-8B-Instruct and GLM4-9B-Chat \cite{glm2024chatglm}) for experimentation. The performance of these models before and after fine-tuning on T-Eval is displayed in Figure \ref{fig:various_llm}. 

It can be seen that for different LLMs, our KG2Tool data is always effective, and the performance improvements of the SFT models are significant. Among the raw models, the performance of Llama3.1-8B and GLM4-9B is inferior to that of Qwen2-7B and Qwen2.5-7B, which may be due to differences in model architecture. Yet, after fine-tuning with our data, their performance gains are very significant, especially for GLM4-9B, which improved from 75.4 to 85.6, surpassing Qwen2.5-7B. This result shows that small models below 10B parameters can perform excellent tool utilization comparable to LLMs like Qwen2.5-72B.

\subsection{Study on General Capabilities}
The results above indicate that LLMs fine-tuned with our constructed instruction data possess superior tool usage capabilities. However, whether their abilities in other aspects are affected also needs further evaluation. To assess the impact of SFT with KG2Tool on the broader capabilities of LLMs, we conduct experiments using Qwen2.5-7B-Instruct across various benchmarks evaluating general ability (MMLU \cite{mmlu}, BBH \cite{suzgun2023bbh}), coding (HumanEval \cite{chen2021humaneval}), mathematics (GSM8K \cite{cobbe2021gsm8k}), and tool utilization (T-Eval).

\begin{figure}[t!]
\centering
\includegraphics[width=0.95\linewidth]{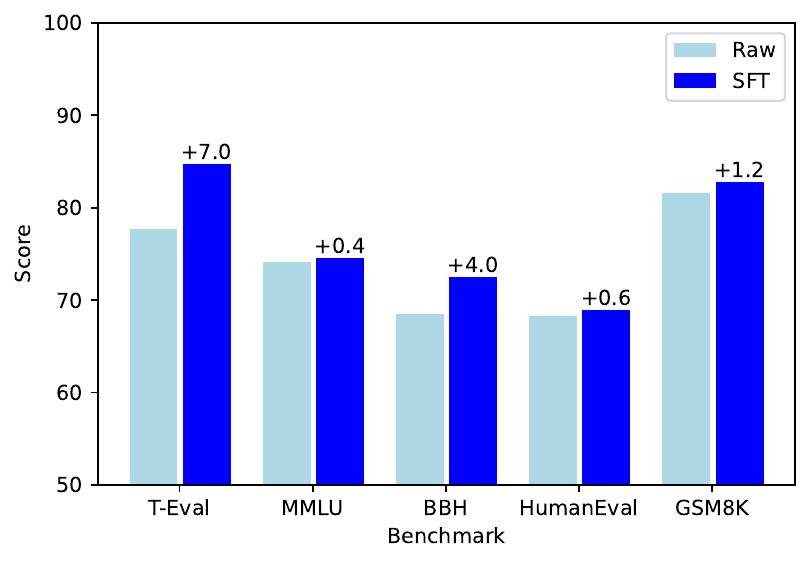}
\caption{General performance of Qwen2.5-7B-Instruct fine-tuned with KG2Tool.}
\label{fig:general_performance}
\end{figure}

Figure \ref{fig:general_performance} presents the performance metrics for both raw and SFT models across these benchmark evaluations. Clearly, the SFT models outperform the RAW models in all benchmarks, indicating that fine-tuning with our data not only fails to weaken the performance of the original model but can even have a positive impact. Notably, the SFT model scored 4.0 points higher than the Raw model on the BBH benchmark, indicating that our data is also significantly effective for the BBH task. Since the BBH task requires multi-step reasoning, it aligns well with how we construct our KG2Tool data. This result suggests that our data can potentially improve the complex reasoning abilities of LLMs.

\section{Conclusion}

This paper proposes a novel data synthesis method that utilizes knowledge graphs to generate high-quality instruction data to enhance the tool utilization performance of LLMs. By leveraging the structural and semantic information of knowledge graphs, we generated API tool sets, queries, and their corresponding solution path sets, thereby constructing the instruction dataset KG2Tool.
Through extensive experiments, we demonstrate that smaller language models fine-tuned with our data can achieve state-of-the-art tool utilization performance, surpassing GPT-4 and LLaMA3.1-70B while maintaining good general capabilities. Our results show that the high quality of data synthesized from KGs can potentially enhance the general capabilities of LLMs.

\section*{Limitations}

We have demonstrated that synthesizing a small amount of data using knowledge graphs can significantly enhance the tool-use capabilities of LLMs. The used knowledge graph should contain many diverse entities and relationships. Small-scale knowledge graphs with limited entities and relationships may not achieve such significant effects.
Furthermore, from the perspective of tool execution efficiency, this paper does not take into account some of the more recently emerged inference models. While it is possible that these models may offer superior tool planning and reasoning capabilities, employing them for tool invocation might not be practical in terms of cost and efficiency. The computational resources required by these advanced models could be substantial, potentially leading to increased latency and higher operational expenses. Additionally, the complexity of integrating such models into existing systems could pose significant challenges, both technically and logistically.



\bibliography{custom}

\begin{thebibliography}{36}
\providecommand{\natexlab}[1]{#1}

\bibitem[{Bach et~al.(2022)Bach, Sanh, Yong, Webson, Raffel, Nayak, Sharma, Kim, Bari, Fevry et~al.}]{bach2022promptsource}
Stephen~H Bach, Victor Sanh, Zheng-Xin Yong, Albert Webson, Colin Raffel, Nihal~V Nayak, Abheesht Sharma, Taewoon Kim, M~Saiful Bari, Thibault Fevry, et~al. 2022.
\newblock Promptsource: An integrated development environment and repository for natural language prompts.
\newblock \emph{arXiv preprint arXiv:2202.01279}.

\bibitem[{Bordes et~al.(2013)Bordes, Usunier, Garcia-Duran, Weston, and Yakhnenko}]{bordes2013translating}
Antoine Bordes, Nicolas Usunier, Alberto Garcia-Duran, Jason Weston, and Oksana Yakhnenko. 2013.
\newblock Translating embeddings for modeling multi-relational data.
\newblock \emph{Advances in neural information processing systems}, 26.

\bibitem[{Chen et~al.(2021)Chen, Tworek, Jun, Yuan, Pinto, Kaplan, Edwards, Burda, Joseph, Brockman et~al.}]{chen2021humaneval}
Mark Chen, Jerry Tworek, Heewoo Jun, Qiming Yuan, Henrique Ponde De~Oliveira Pinto, Jared Kaplan, Harri Edwards, Yuri Burda, Nicholas Joseph, Greg Brockman, et~al. 2021.
\newblock Evaluating large language models trained on code.
\newblock \emph{arXiv preprint arXiv:2107.03374}.

\bibitem[{Chen et~al.(2023)Chen, Du, Zhang, Liu, Liu, Zheng, Zhuo, Zhang, Lin, Chen et~al.}]{chen2023teval}
Zehui Chen, Weihua Du, Wenwei Zhang, Kuikun Liu, Jiangning Liu, Miao Zheng, Jingming Zhuo, Songyang Zhang, Dahua Lin, Kai Chen, et~al. 2023.
\newblock T-eval: Evaluating the tool utilization capability step by step.
\newblock \emph{arXiv preprint arXiv:2312.14033}.

\bibitem[{Chen et~al.(2024)Chen, Liu, Wang, Zhang, Liu, Lin, Chen, and Zhao}]{chen2024agentflan}
Zehui Chen, Kuikun Liu, Qiuchen Wang, Wenwei Zhang, Jiangning Liu, Dahua Lin, Kai Chen, and Feng Zhao. 2024.
\newblock Agent-flan: Designing data and methods of effective agent tuning for large language models.
\newblock \emph{arXiv preprint arXiv:2403.12881}.

\bibitem[{Cobbe et~al.(2021)Cobbe, Kosaraju, Bavarian, Chen, Jun, Kaiser, Plappert, Tworek, Hilton, Nakano et~al.}]{cobbe2021gsm8k}
Karl Cobbe, Vineet Kosaraju, Mohammad Bavarian, Mark Chen, Heewoo Jun, Lukasz Kaiser, Matthias Plappert, Jerry Tworek, Jacob Hilton, Reiichiro Nakano, et~al. 2021.
\newblock Training verifiers to solve math word problems.
\newblock \emph{arXiv preprint arXiv:2110.14168}.

\bibitem[{Dubey et~al.(2024)Dubey, Jauhri, Pandey, Kadian, Al-Dahle, Letman, Mathur, Schelten, Yang, Fan et~al.}]{dubey2024llama}
Abhimanyu Dubey, Abhinav Jauhri, Abhinav Pandey, Abhishek Kadian, Ahmad Al-Dahle, Aiesha Letman, Akhil Mathur, Alan Schelten, Amy Yang, Angela Fan, et~al. 2024.
\newblock The llama 3 herd of models.
\newblock \emph{arXiv preprint arXiv:2407.21783}.

\bibitem[{Gao et~al.(2023)Gao, Madaan, Zhou, Alon, Liu, Yang, Callan, and Neubig}]{gao2023pal}
Luyu Gao, Aman Madaan, Shuyan Zhou, Uri Alon, Pengfei Liu, Yiming Yang, Jamie Callan, and Graham Neubig. 2023.
\newblock Pal: Program-aided language models.
\newblock In \emph{International Conference on Machine Learning}, pages 10764--10799. PMLR.

\bibitem[{GLM et~al.(2024)GLM, Zeng, Xu, Wang, Zhang, Yin, Rojas, Feng, Zhao, Lai et~al.}]{glm2024chatglm}
Team GLM, Aohan Zeng, Bin Xu, Bowen Wang, Chenhui Zhang, Da~Yin, Diego Rojas, Guanyu Feng, Hanlin Zhao, Hanyu Lai, et~al. 2024.
\newblock Chatglm: A family of large language models from glm-130b to glm-4 all tools.
\newblock \emph{arXiv preprint arXiv:2406.12793}.

\bibitem[{Gunasekar et~al.(2023)Gunasekar, Zhang, Aneja, Mendes, Del~Giorno, Gopi, Javaheripi, Kauffmann, de~Rosa, Saarikivi et~al.}]{gunasekar2023textbooks}
Suriya Gunasekar, Yi~Zhang, Jyoti Aneja, Caio C{\'e}sar~Teodoro Mendes, Allie Del~Giorno, Sivakanth Gopi, Mojan Javaheripi, Piero Kauffmann, Gustavo de~Rosa, Olli Saarikivi, et~al. 2023.
\newblock Textbooks are all you need.
\newblock \emph{arXiv preprint arXiv:2306.11644}.

\bibitem[{Hendrycks et~al.(2021)Hendrycks, Burns, Basart, Zou, Mazeika, Song, and Steinhardt}]{mmlu}
Dan Hendrycks, Collin Burns, Steven Basart, Andy Zou, Mantas Mazeika, Dawn Song, and Jacob Steinhardt. 2021.
\newblock Measuring massive multitask language understanding.
\newblock In \emph{International Conference on Learning Representations}.

\bibitem[{Hu et~al.(2021)Hu, Shen, Wallis, Allen-Zhu, Li, Wang, Wang, and Chen}]{hu2021lora}
Edward~J Hu, Yelong Shen, Phillip Wallis, Zeyuan Allen-Zhu, Yuanzhi Li, Shean Wang, Lu~Wang, and Weizhu Chen. 2021.
\newblock Lora: Low-rank adaptation of large language models.
\newblock \emph{arXiv preprint arXiv:2106.09685}.

\bibitem[{Jiang et~al.(2023)Jiang, Zhou, Dong, Ye, Zhao, and Wen}]{jiang2023structgpt}
Jinhao Jiang, Kun Zhou, Zican Dong, Keming Ye, Wayne~Xin Zhao, and Ji-Rong Wen. 2023.
\newblock Structgpt: A general framework for large language model to reason over structured data.
\newblock \emph{arXiv preprint arXiv:2305.09645}.

\bibitem[{Kadlec et~al.(2017)Kadlec, Bajgar, and Kleindienst}]{kadlec2017knowledge}
Rudolf Kadlec, Ondrej Bajgar, and Jan Kleindienst. 2017.
\newblock Knowledge base completion: Baselines strike back.
\newblock \emph{arXiv preprint arXiv:1705.10744}.

\bibitem[{Khattab et~al.(2022)Khattab, Santhanam, Li, Hall, Liang, Potts, and Zaharia}]{khattab2022demonstrate}
Omar Khattab, Keshav Santhanam, Xiang~Lisa Li, David Hall, Percy Liang, Christopher Potts, and Matei Zaharia. 2022.
\newblock Demonstrate-search-predict: Composing retrieval and language models for knowledge-intensive nlp.
\newblock \emph{arXiv preprint arXiv:2212.14024}.

\bibitem[{Kwon et~al.(2023)Kwon, Li, Zhuang, Sheng, Zheng, Yu, Gonzalez, Zhang, and Stoica}]{kwon2023vllm}
Woosuk Kwon, Zhuohan Li, Siyuan Zhuang, Ying Sheng, Lianmin Zheng, Cody~Hao Yu, Joseph~E. Gonzalez, Hao Zhang, and Ion Stoica. 2023.
\newblock Efficient memory management for large language model serving with pagedattention.
\newblock In \emph{Proceedings of the ACM SIGOPS 29th Symposium on Operating Systems Principles}.

\bibitem[{Lewis et~al.(2020)Lewis, Perez, Piktus, Petroni, Karpukhin, Goyal, K{\"u}ttler, Lewis, Yih, Rockt{\"a}schel et~al.}]{lewis2020retrieval}
Patrick Lewis, Ethan Perez, Aleksandra Piktus, Fabio Petroni, Vladimir Karpukhin, Naman Goyal, Heinrich K{\"u}ttler, Mike Lewis, Wen-tau Yih, Tim Rockt{\"a}schel, et~al. 2020.
\newblock Retrieval-augmented generation for knowledge-intensive nlp tasks.
\newblock \emph{Advances in Neural Information Processing Systems}, 33:9459--9474.

\bibitem[{Liu et~al.(2024)Liu, Huang, Zeng, Hao, Yu, Li, Wang, Gan, Liu, Yu et~al.}]{liu2024toolace}
Weiwen Liu, Xu~Huang, Xingshan Zeng, Xinlong Hao, Shuai Yu, Dexun Li, Shuai Wang, Weinan Gan, Zhengying Liu, Yuanqing Yu, et~al. 2024.
\newblock Toolace: Winning the points of llm function calling.
\newblock \emph{arXiv preprint arXiv:2409.00920}.

\bibitem[{Lu et~al.(2024)Lu, Peng, Cheng, Galley, Chang, Wu, Zhu, and Gao}]{lu2024chameleon}
Pan Lu, Baolin Peng, Hao Cheng, Michel Galley, Kai-Wei Chang, Ying~Nian Wu, Song-Chun Zhu, and Jianfeng Gao. 2024.
\newblock Chameleon: Plug-and-play compositional reasoning with large language models.
\newblock \emph{Advances in Neural Information Processing Systems}, 36.

\bibitem[{Luo et~al.(2023)Luo, Li, Haffari, and Pan}]{luo2023reasoning}
Linhao Luo, Yuan-Fang Li, Gholamreza Haffari, and Shirui Pan. 2023.
\newblock Reasoning on graphs: Faithful and interpretable large language model reasoning.
\newblock \emph{arXiv preprint arXiv:2310.01061}.

\bibitem[{Qin et~al.(2024)Qin, Hu, Lin, Chen, Ding, Cui, Zeng, Huang, Xiao, Han, Fung, Su, Wang, Qian, Tian, Zhu, Liang, Shen, Xu, Zhang, Ye, Li, Tang, Yi, Zhu, Dai, Yan, Cong, Lu, Zhao, Huang, Yan, Han, Sun, Li, Phang, Yang, Wu, Ji, Liu, and Sun}]{qin2024toollearningfoundationmodels}
Yujia Qin, Shengding Hu, Yankai Lin, Weize Chen, Ning Ding, Ganqu Cui, Zheni Zeng, Yufei Huang, Chaojun Xiao, Chi Han, Yi~Ren Fung, Yusheng Su, Huadong Wang, Cheng Qian, Runchu Tian, Kunlun Zhu, Shihao Liang, Xingyu Shen, Bokai Xu, Zhen Zhang, Yining Ye, Bowen Li, Ziwei Tang, Jing Yi, Yuzhang Zhu, Zhenning Dai, Lan Yan, Xin Cong, Yaxi Lu, Weilin Zhao, Yuxiang Huang, Junxi Yan, Xu~Han, Xian Sun, Dahai Li, Jason Phang, Cheng Yang, Tongshuang Wu, Heng Ji, Zhiyuan Liu, and Maosong Sun. 2024.
\newblock \href {https://arxiv.org/abs/2304.08354} {Tool learning with foundation models}.
\newblock \emph{Preprint}, arXiv:2304.08354.

\bibitem[{Qin et~al.(2023)Qin, Liang, Ye, Zhu, Yan, Lu, Lin, Cong, Tang, Qian et~al.}]{qin2023toolllm}
Yujia Qin, Shihao Liang, Yining Ye, Kunlun Zhu, Lan Yan, Yaxi Lu, Yankai Lin, Xin Cong, Xiangru Tang, Bill Qian, et~al. 2023.
\newblock Toolllm: Facilitating large language models to master 16000+ real-world apis.
\newblock \emph{arXiv preprint arXiv:2307.16789}.

\bibitem[{Schick et~al.(2024)Schick, Dwivedi-Yu, Dess{\`\i}, Raileanu, Lomeli, Hambro, Zettlemoyer, Cancedda, and Scialom}]{schick2024toolformer}
Timo Schick, Jane Dwivedi-Yu, Roberto Dess{\`\i}, Roberta Raileanu, Maria Lomeli, Eric Hambro, Luke Zettlemoyer, Nicola Cancedda, and Thomas Scialom. 2024.
\newblock Toolformer: Language models can teach themselves to use tools.
\newblock \emph{Advances in Neural Information Processing Systems}, 36.

\bibitem[{Shen et~al.(2024)Shen, Song, Tan, Li, Lu, and Zhuang}]{shen2024hugginggpt}
Yongliang Shen, Kaitao Song, Xu~Tan, Dongsheng Li, Weiming Lu, and Yueting Zhuang. 2024.
\newblock Hugginggpt: Solving ai tasks with chatgpt and its friends in hugging face.
\newblock \emph{Advances in Neural Information Processing Systems}, 36.

\bibitem[{Srinivasan et~al.(2023)Srinivasan, Dong, Zhu, Yu, Mosk-Aoyama, Keutzer, Jiao, and Zhang}]{srinivasan2023nexusraven}
Venkat~Krishna Srinivasan, Zhen Dong, Banghua Zhu, Brian Yu, Damon Mosk-Aoyama, Kurt Keutzer, Jiantao Jiao, and Jian Zhang. 2023.
\newblock Nexusraven: a commercially-permissive language model for function calling.
\newblock In \emph{NeurIPS 2023 Foundation Models for Decision Making Workshop}.

\bibitem[{Suzgun et~al.(2023)Suzgun, Scales, Sch{\"a}rli, Gehrmann, Tay, Chung, Chowdhery, Le, Chi, Zhou et~al.}]{suzgun2023bbh}
Mirac Suzgun, Nathan Scales, Nathanael Sch{\"a}rli, Sebastian Gehrmann, Yi~Tay, Hyung~Won Chung, Aakanksha Chowdhery, Quoc Le, Ed~Chi, Denny Zhou, et~al. 2023.
\newblock Challenging big-bench tasks and whether chain-of-thought can solve them.
\newblock In \emph{Findings of the Association for Computational Linguistics: ACL 2023}, pages 13003--13051.

\bibitem[{Tang et~al.(2023)Tang, Deng, Lin, Han, Liang, Cao, and Sun}]{tang2023toolalpaca}
Qiaoyu Tang, Ziliang Deng, Hongyu Lin, Xianpei Han, Qiao Liang, Boxi Cao, and Le~Sun. 2023.
\newblock Toolalpaca: Generalized tool learning for language models with 3000 simulated cases.
\newblock \emph{arXiv preprint arXiv:2306.05301}.

\bibitem[{Team(2024)}]{qwen2.5}
Qwen Team. 2024.
\newblock \href {https://qwenlm.github.io/blog/qwen2.5/} {Qwen2.5: A party of foundation models}.

\bibitem[{Wang et~al.(2024)Wang, Chen, Hu, Yang, Liu, Shen, Wei, Zhang, Gu, Zhou et~al.}]{wang2024learning}
Junjie Wang, Mingyang Chen, Binbin Hu, Dan Yang, Ziqi Liu, Yue Shen, Peng Wei, Zhiqiang Zhang, Jinjie Gu, Jun Zhou, et~al. 2024.
\newblock Learning to plan for retrieval-augmented large language models from knowledge graphs.
\newblock \emph{arXiv preprint arXiv:2406.14282}.

\bibitem[{Wei et~al.(2021)Wei, Bosma, Zhao, Guu, Yu, Lester, Du, Dai, and Le}]{wei2021finetuned}
Jason Wei, Maarten Bosma, Vincent~Y Zhao, Kelvin Guu, Adams~Wei Yu, Brian Lester, Nan Du, Andrew~M Dai, and Quoc~V Le. 2021.
\newblock Finetuned language models are zero-shot learners.
\newblock \emph{arXiv preprint arXiv:2109.01652}.

\bibitem[{Yao et~al.(2022)Yao, Zhao, Yu, Du, Shafran, Narasimhan, and Cao}]{yao2022react}
Shunyu Yao, Jeffrey Zhao, Dian Yu, Nan Du, Izhak Shafran, Karthik Narasimhan, and Yuan Cao. 2022.
\newblock React: Synergizing reasoning and acting in language models.
\newblock \emph{arXiv preprint arXiv:2210.03629}.

\bibitem[{Ye et~al.(2025)Ye, Du, Yao, Lin, Xu, Chen, Wang, Zhu, Xi, Yuan, Gui, Zhang, Huang, and Chen}]{Ye2025ToolHop}
Junjie Ye, Zhengyin Du, Xuesong Yao, Weijian Lin, Yufei Xu, Zehui Chen, Zaiyuan Wang, Sining Zhu, Zhiheng Xi, Siyu Yuan, Tao Gui, Qi~Zhang, Xuanjing Huang, and Jiecao Chen. 2025.
\newblock \href {https://doi.org/10.48550/ARXIV.2501.02506} {Toolhop: {A} query-driven benchmark for evaluating large language models in multi-hop tool use}.
\newblock \emph{CoRR}, abs/2501.02506.

\bibitem[{Yuan et~al.(2023)Yuan, Chen, Wang, Fung, Peng, and Ji}]{yuan2023craft}
Lifan Yuan, Yangyi Chen, Xingyao Wang, Yi~R Fung, Hao Peng, and Heng Ji. 2023.
\newblock Craft: Customizing llms by creating and retrieving from specialized toolsets.
\newblock \emph{arXiv preprint arXiv:2309.17428}.

\bibitem[{Zhou et~al.(2024{\natexlab{a}})Zhou, Liu, Xu, Iyer, Sun, Mao, Ma, Efrat, Yu, Yu et~al.}]{zhou2024lima}
Chunting Zhou, Pengfei Liu, Puxin Xu, Srinivasan Iyer, Jiao Sun, Yuning Mao, Xuezhe Ma, Avia Efrat, Ping Yu, Lili Yu, et~al. 2024{\natexlab{a}}.
\newblock Lima: Less is more for alignment.
\newblock \emph{Advances in Neural Information Processing Systems}, 36.

\bibitem[{Zhou et~al.(2024{\natexlab{b}})Zhou, Ghaddar, Zhang, Ma, Hu, Pal, Coates, Wang, Zhang, and Hao}]{zhou2024enhancing}
Jiaming Zhou, Abbas Ghaddar, Ge~Zhang, Liheng Ma, Yaochen Hu, Soumyasundar Pal, Mark Coates, Bin Wang, Yingxue Zhang, and Jianye Hao. 2024{\natexlab{b}}.
\newblock Enhancing logical reasoning in large language models through graph-based synthetic data.
\newblock \emph{arXiv preprint arXiv:2409.12437}.

\bibitem[{Zhu et~al.(2022)Zhu, Galkin, Zhang, and Tang}]{zhu2022neural}
Zhaocheng Zhu, Mikhail Galkin, Zuobai Zhang, and Jian Tang. 2022.
\newblock Neural-symbolic models for logical queries on knowledge graphs.
\newblock In \emph{International conference on machine learning}, pages 27454--27478. PMLR.

\end{thebibliography}

\appendix

\section{Appendix}
\label{sec:appendix}

\subsection{FOL Query Patterns}
Tables \ref{tab:pattern_part1} and \ref{tab:pattern_part2} depict all 14 FOL query patterns and their corresponding FOL Subgraphs in detail. In each row, we provide an instantiated query example in FOL form and its translated query in natural language form for each FOL query pattern.

\subsection{Prompts for Generating Data}
\begin{figure*}[t!]
\centering
\includegraphics[width=\linewidth]{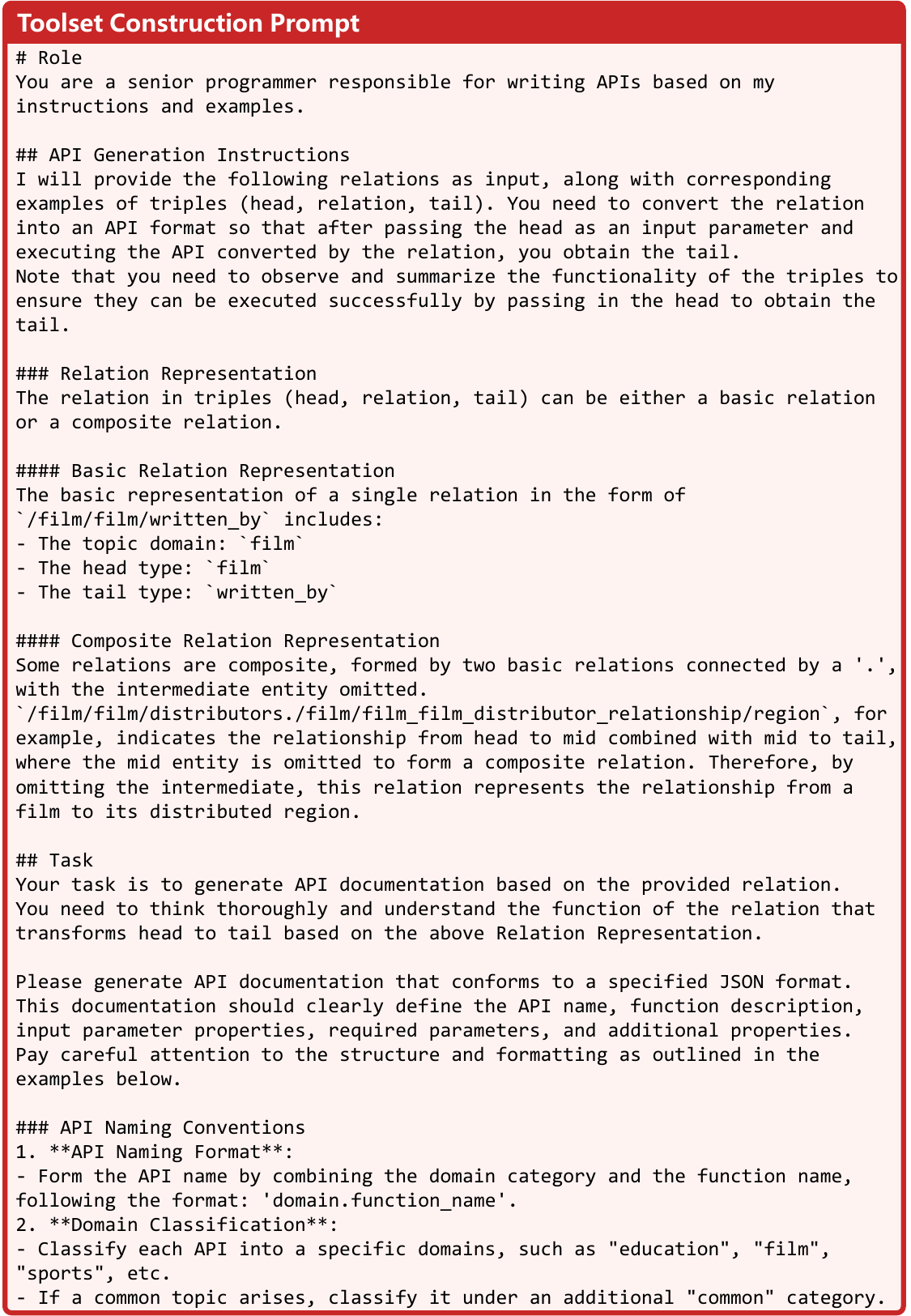}
\caption{Prompt for API Generation using Relations (Part1).}
\label{fig:rel2api_p1}
\end{figure*}

\begin{figure*}[t!]
\centering
\includegraphics[width=\linewidth]{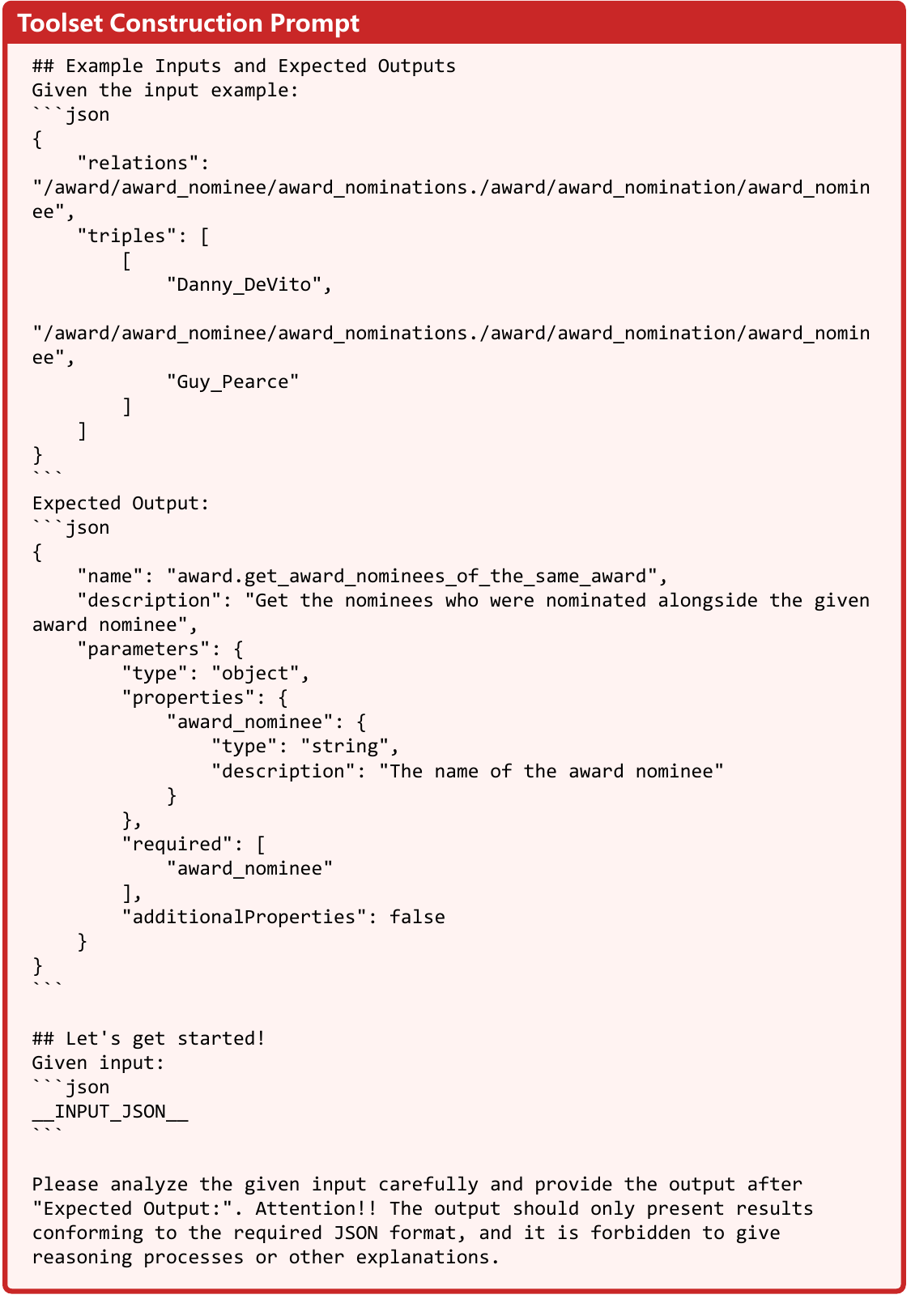}
\caption{Prompt for API Generation using Relations (Part2).}
\label{fig:rel2api_p2}
\end{figure*}

\begin{figure*}[t!]
\centering
\includegraphics[width=\linewidth]{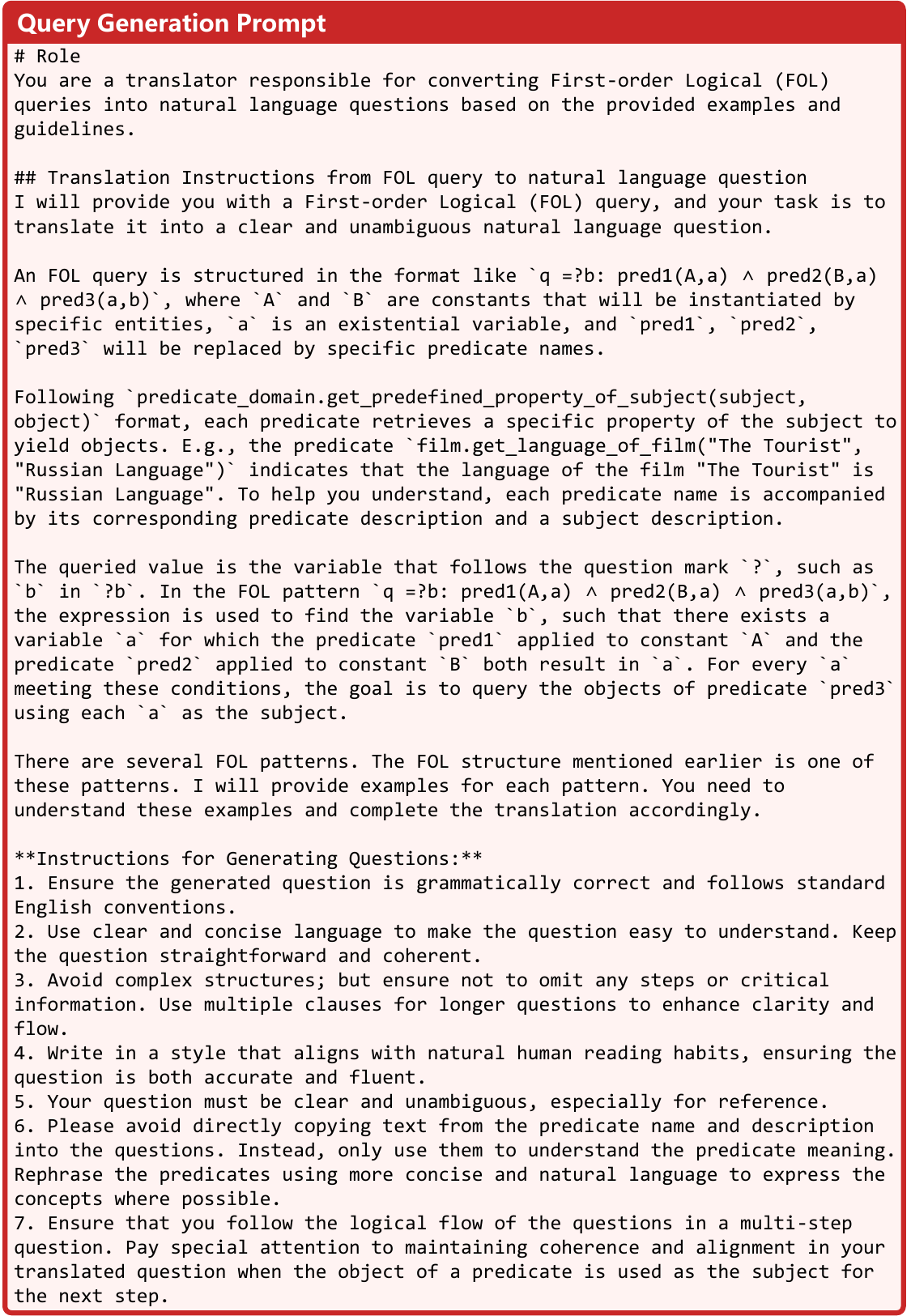}
\caption{Prompt for Translating Instantiated FOL Queries into Natural Language Form (Part1).}
\label{fig:fol2query_p1}
\end{figure*}

\begin{figure*}[t!]
\centering
\includegraphics[width=\linewidth]{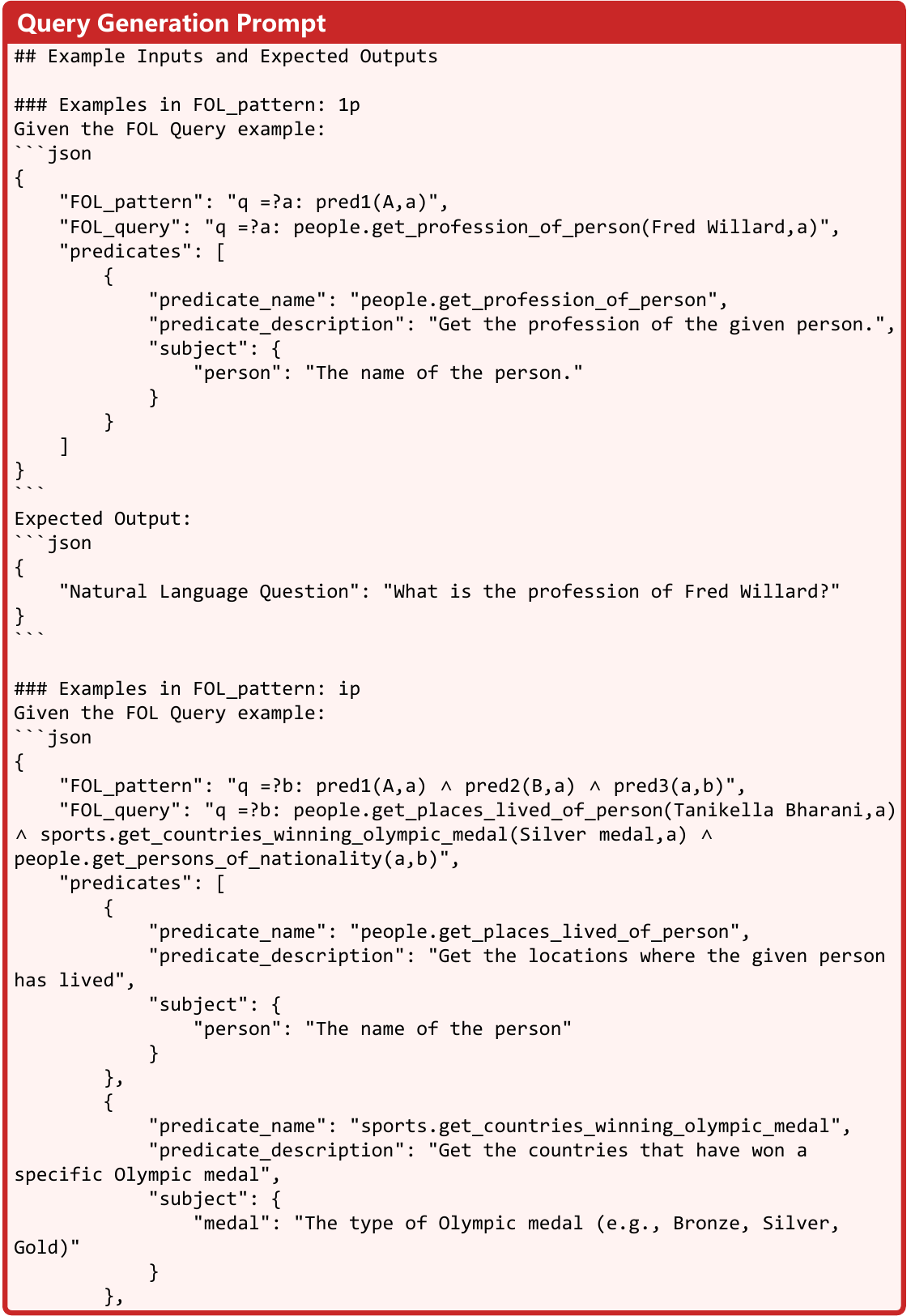}
\caption{Prompt for Translating Instantiated FOL Queries into Natural Language Form (Part2).}
\label{fig:fol2query_p2}
\end{figure*}

\begin{figure*}[t!]
\centering
\includegraphics[width=\linewidth]{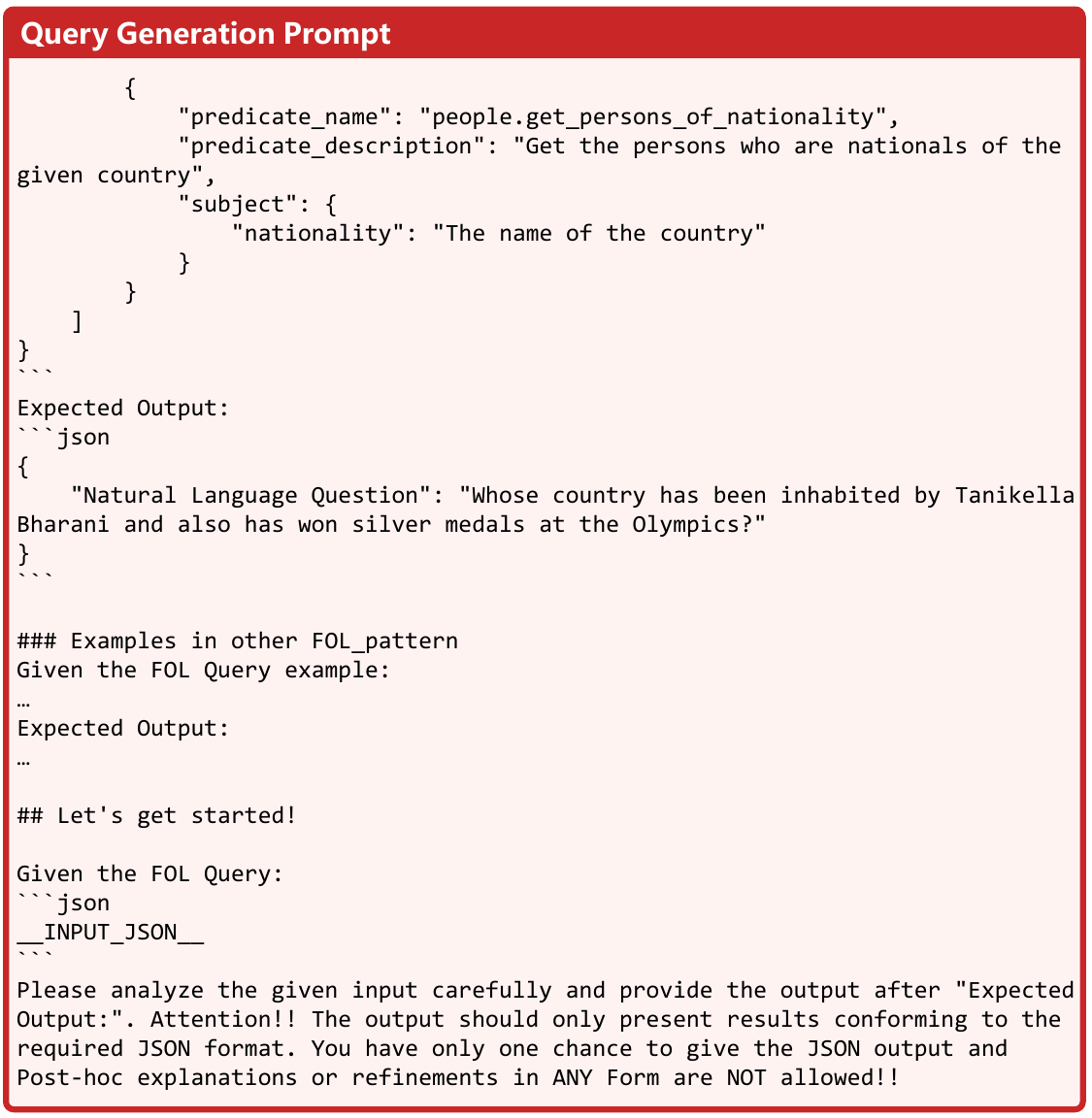}
\caption{Prompt for Translating Instantiated FOL Queries into Natural Language Form (Part3).}
\label{fig:fol2query_p3}
\end{figure*}

Figures \ref{fig:rel2api_p1} and \ref{fig:rel2api_p2} show the prompt for API generation using triple relations in knowledge graphs.
Figures \ref{fig:fol2query_p1}, \ref{fig:fol2query_p2}, and \ref{fig:fol2query_p3} show the prompt to translate an instantiated FOL query of a specific pattern into its natural language form.
We use a single prompt to convert FOL queries of all patterns into natural language questions. This prompt includes typical examples for each pattern for the LLM to reference. 
Since the relation in KG is represented in the URL format, which is often difficult to interpret, we replace the original relations with the API names generated in the previous step. With the input parameters and descriptions of the APIs alongside the FOL query, LLMs fully understand the semantics of the relations and focus on the query translation task itself.

It is worth noting that, due to space constraints and to avoid redundancy, only snippets of the conversion examples for \textit{1p} and \textit{ip} patterns are presented here. 
The full prompt can be accessed in our later-released project repository after review.

\subsection{Details of Constructed Instruction Data}
\label{a3}
\begin{figure*}[t!]
\centering
\includegraphics[width=\linewidth]{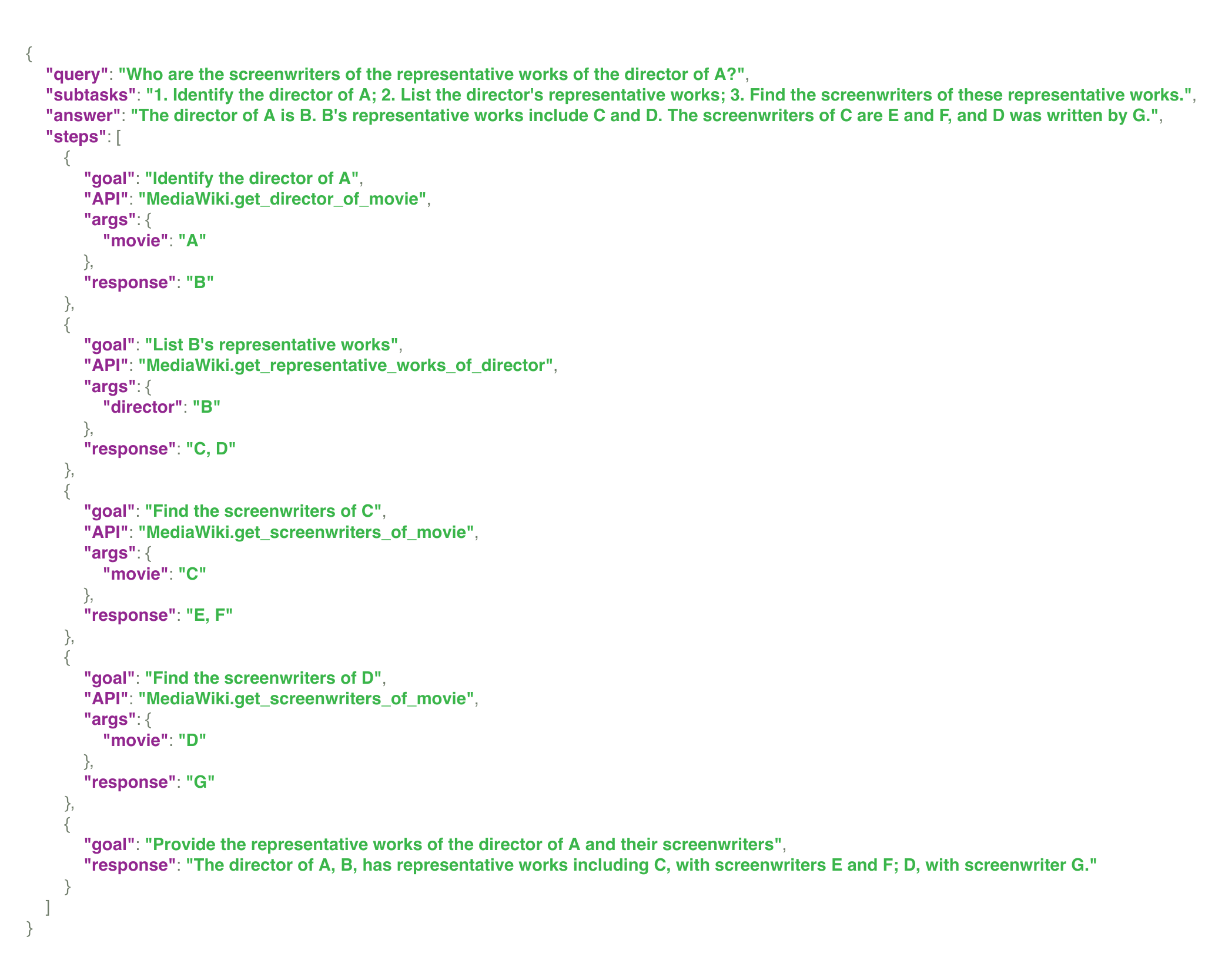}
\caption{Example of a demo query-solution pair in the $3p$ pattern.}
\label{fig:demo_qs}
\end{figure*}

For each FOL query pattern, we generated thousands of complete query-solution pairs from the KG. As mentioned in the main text, each solution includes detailed steps to solve the specified query. Figure \ref{fig:demo_qs} shows an example of a demo query-solution pair in the $3p$ pattern, including a query in natural language, the decomposed subtasks, the final answer, and the detailed steps to solve each subtask. Note that each step includes complete information such as the goal, the API used, the input arguments, and the tool's response.

\begin{figure*}[t!]
\centering
\includegraphics[width=\linewidth]{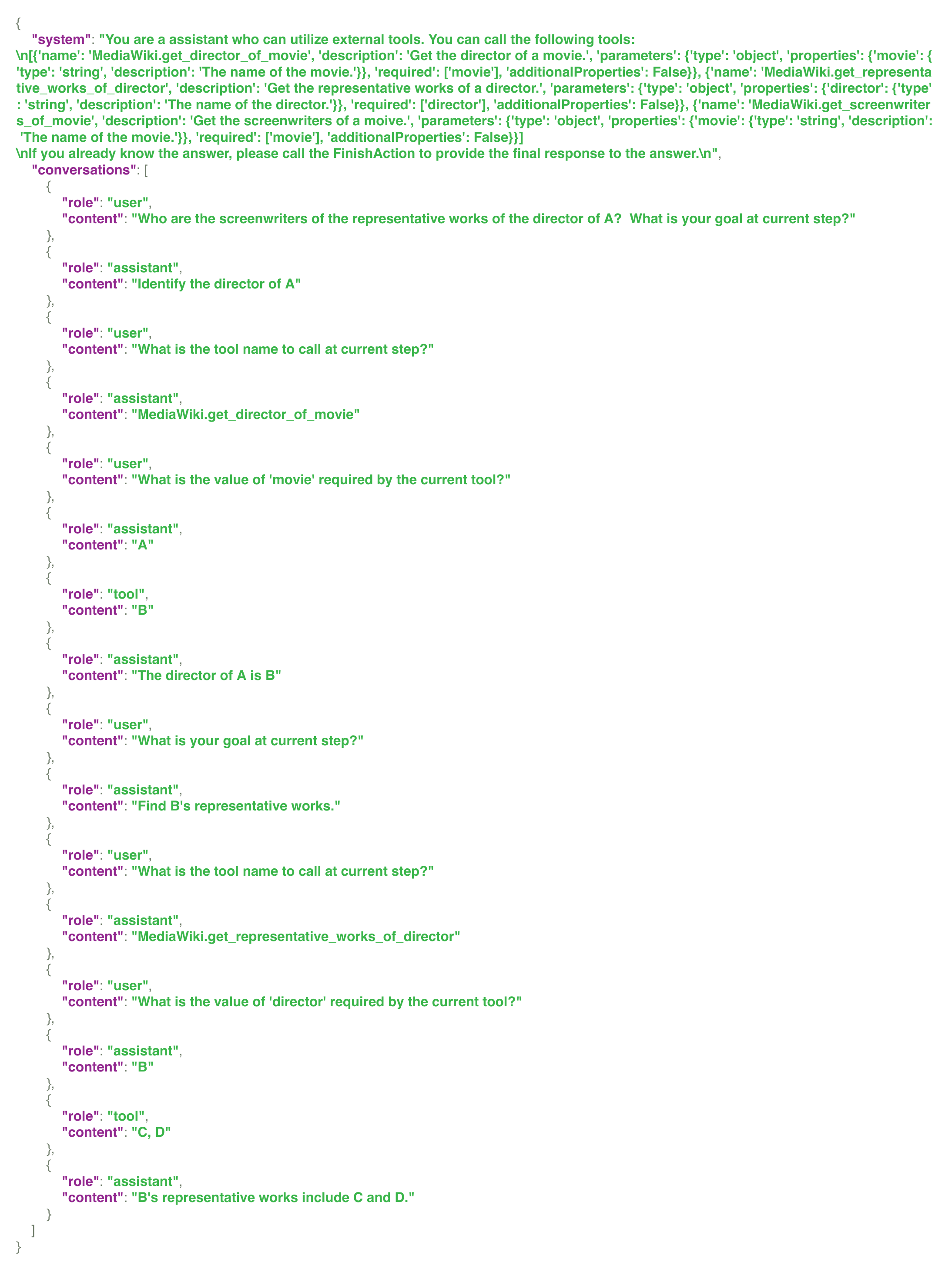}
\caption{Example of instruction data formatted as a dialogue between the user and the assistant.}
\label{fig:demo_chat}
\end{figure*}

Then, we iterate through each step in every query-solution pair. For each step, we construct question-answer pairs. The questions can be about the goal of the current step or which tool should be called in the current step. We also design planning-type questions for the initial query, requiring the model to break down the problem and provide subtasks. Additionally, we replace some real tool responses with fake or incorrect ones and then ask the model to determine whether the tool response solves the problem. This operation generates some review-type question-answer pairs. Figure \ref{fig:demo_chat} shows an example of instruction data formatted as a dialogue between the user and the assistant.

Through the above operations, we can quickly construct tens of thousands of instruction samples for each FOL query pattern. Specifically, we randomly sample 1,000 instances from the data of each FOL pattern, and by combining all these samples together, we obtain the KG2Tool dataset. Finally, we randomly sample a small subset of instances from the KG2Tool for fine-tuning the LLMs.

\begin{table*}[t] 
\caption{Illustration of FOL Query Patterns and FOL Subgraphs with Query Examples in FOL and Natural Language Form (Part 1)}
\label{tab:pattern_part1}
\centering
\renewcommand{\arraystretch}{1.3}
\begin{tabular}{
     >{\centering\arraybackslash}m{0.8cm}
    | >{\centering\arraybackslash}m{1.7cm}
    | >{\centering\arraybackslash}m{3cm}
    | >{\centering\arraybackslash}m{4cm}
    | >{\centering\arraybackslash}m{4.5cm} 
}
\toprule
Query Type & FOL Subgraph & FOL Query Pattern & Instantiated Query Example in FOL Form & Query Example in Natural Language Form\\
\midrule
1p   & \includegraphics[width=2cm]{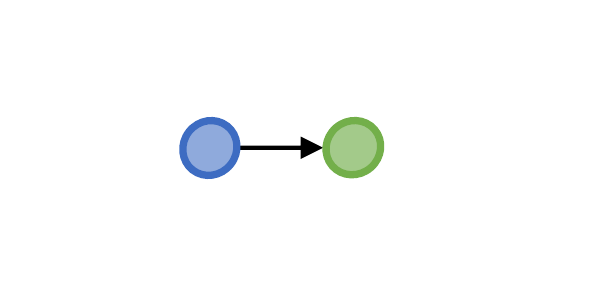} & q =?a : Rel\_1(A, a) & q =?a : Star(Amy Irving, a) & What film did Amy Irving star in? \\ \midrule
2p   & \includegraphics[width=2cm]{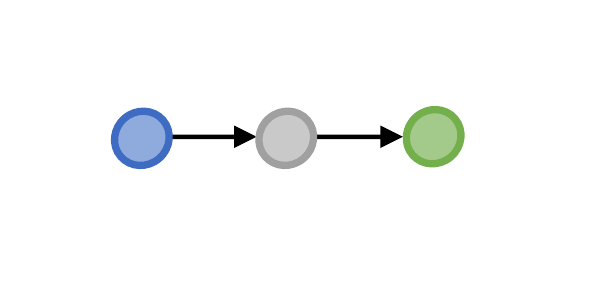} & q =?b : Rel\_1(A, a) $\wedge$ Rel\_2(a, b) & q =?b : Nominated(Lee Grant, a) $\wedge$ Winner(a, b) & Who is the winner of the award that Lee Grant was nominated for? \\ \midrule
3p & \includegraphics[width=2cm]{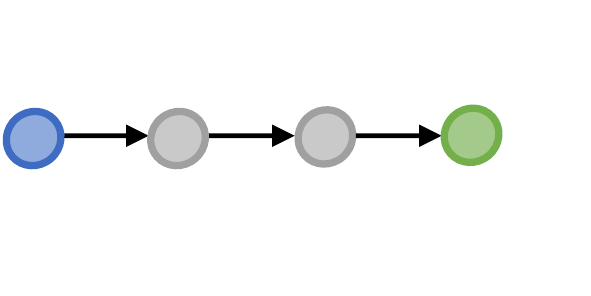} & q =?c : $\exists$ a, b : Rel\_1(A, a) $\wedge$ Rel\_2(a, b) $\wedge$ Rel\_3(b, c) & q =?c : $\exists$ a, b : MemberStates(WTO, a) $\wedge$ JurisdicationOfOffice(b, a) $\wedge$ Organization(b, c) & What is the organization that a politician of a WTO member state came from? \\  \midrule
2i & \includegraphics[width=2cm]{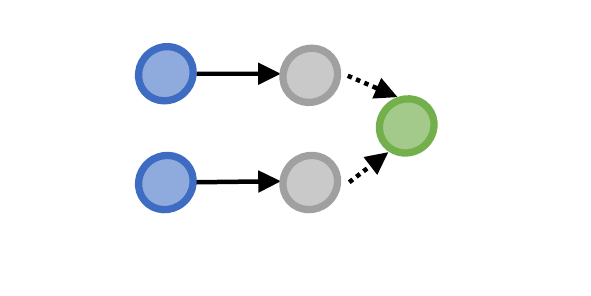} & q =?c : Rel\_1(A, c) $\wedge$ Rel\_2(B, c) & q =?c : Genre(c, Science Fiction) $\wedge$ DistributeFilm(Warner Bros., c) & Which science fiction film is distributed by Warner Bros.? \\ \midrule
3i & \includegraphics[width=2cm]{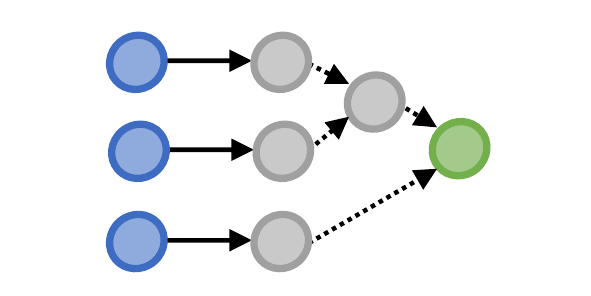} & q =?e : Rel\_1(A, e) $\wedge$ Rel\_2(B, e) $\wedge$ Rel\_3(C, e) & q =?e : Profession(e, songwriter) $\wedge$ WinnerOfSameAward(e, BeBe Winans) $\wedge$ NominatedForSameAward(Babyface, e) & Which songwriter won the same award as BeBeWinans, and was nominated for the same award as Babyface? \\ \midrule
pi & \includegraphics[width=2cm]{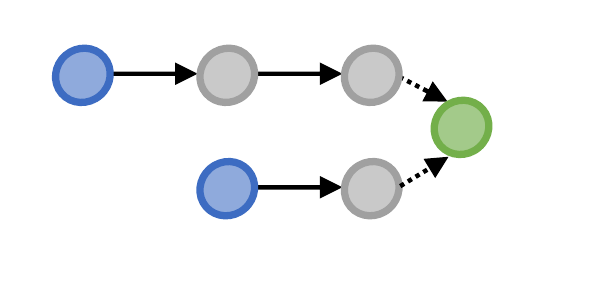} & q =?d : $\exists$ a : Rel\_1(A, a) $\wedge$ Rel\_2(a, d) $\wedge$ Rel\_3(B, d) & q =?d : $\exists$ a : Student(West Point, a) $\wedge$ Found(a, d) $\wedge$ Company(Buzz Aldrin, d) & What is contained by both the administrative division of Columbia and South Carolina? \\ \midrule
ip & \includegraphics[width=2cm]{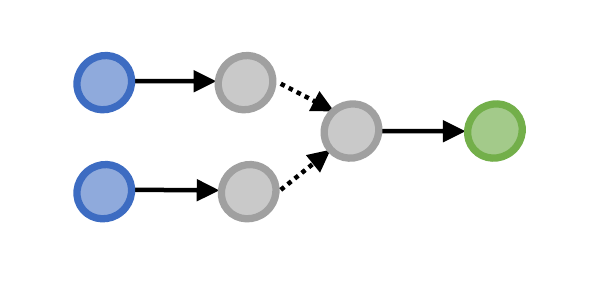} & q =?d : $\exists$ c : Rel\_1(A, c) $\wedge$ Rel\_2(B, c) $\wedge$ Rel\_3(c, d) & q =?d : $\exists$ c : Award(Freddy Got Fingered, c) $\wedge$ Nominated(Peter Hyams, c) $\wedge$ NominatedFor(d, c) & What film is nominated for the award that Freddy Got Fingered won and Peter Hyams was nominated for? \\ \midrule
2u & \includegraphics[width=2cm]{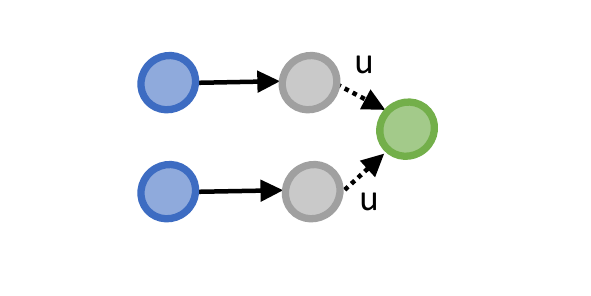} & q =?c : Rel\_1(A, c) $\vee$ Rel\_2(B, c) & q =?c : Student(Bucknell University, c) $\vee$ Nominated(c, National Book Award for Fiction) & Who is a student of Bucknell University or is nominated for National Book Award for Fiction? \\ \midrule
up & \includegraphics[width=2cm]{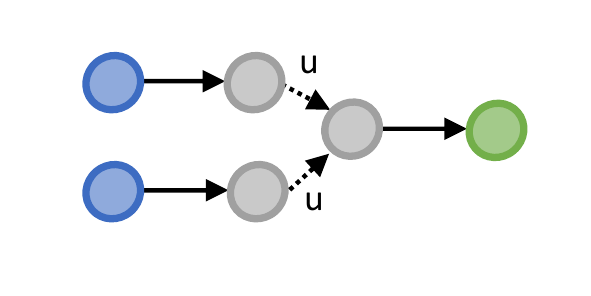} & q =?d : $\exists$ c : (Rel\_1(A, c) $\vee$ Rel\_2(B, c)) $\wedge$ Rel\_3(c, d) & q =?d : $\exists$ c : (Award(David Kirschner, c) $\vee$ Ceremony(c, 39th Daytime Emmy Awards)) $\wedge$ Award(d, c) & Who wins the award that David Kirschner winned or is given at 39th Daytime Emmy Awards? \\ 
\bottomrule
\end{tabular}
\end{table*}

\clearpage 
\onecolumn
\noindent 
\vspace*{0pt} 
\begin{table*}[t] 
\caption{Illustration of FOL Query Patterns and FOL Subgraphs with Query Examples in FOL and Natural Language Form (Part 2)}
\label{tab:pattern_part2}
\centering
\renewcommand{\arraystretch}{1.3}
\begin{tabular}{
     >{\centering\arraybackslash}m{0.8cm}
    | >{\centering\arraybackslash}m{1.7cm}
    | >{\centering\arraybackslash}m{3cm}
    | >{\centering\arraybackslash}m{4cm}
    | >{\centering\arraybackslash}m{4.5cm} 
}
\toprule
Query Type & FOL Subgraph & FOL Query Pattern & Instantiated Query Example in FOL Form & Query Example in Natural Language Form\\
\midrule
2in & \includegraphics[width=2cm]{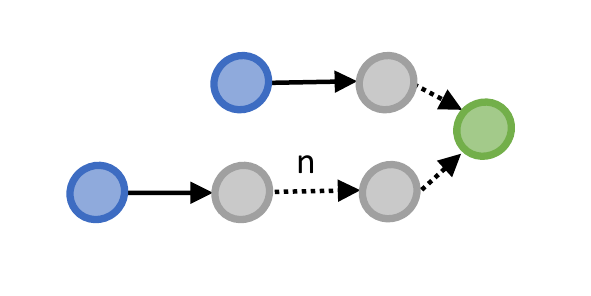} & q =?d : Rel\_1(A, d) $\wedge$ $\neg$Rel\_2(B, d)  & q =?d : MortgageSource(d, US Department of HUD) $\wedge$ $\neg$PlaceLive(Allison Janney, d) & Which city takes mortgage from US Department of Housing and Urban Development, but Allison Janney hasn’t lived in? \\ \midrule
3in & \includegraphics[width=2cm]{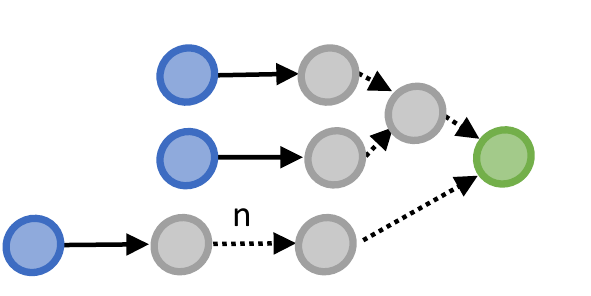} & q =?f : Rel\_1(A, f) $\wedge$ Rel\_2(B, f) $\wedge$ $\neg$Rel\_3(C, f) & q =?f : ReleaseMedium(f, DVD) $\wedge$ ReleaseRegion(f, New Zealand) $\wedge$ $\neg$Language(f, English) & Which film has a DVD version and is released in New Zealand, but doesn’t have an English version? \\ \midrule
inp & \includegraphics[width=2cm]{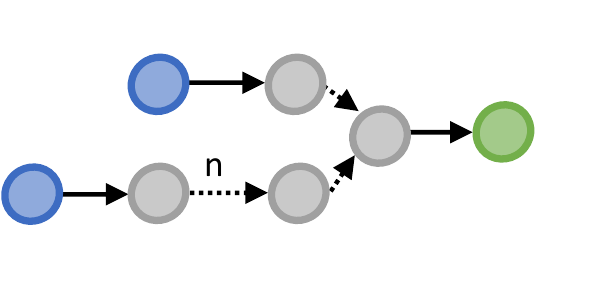} & q =?e : $\exists$ d : Rel\_1(A, d) $\wedge$ $\neg$Rel\_2(B, d) $\wedge$ Rel\_1(d, e) & q =?e : $\exists$ d : FieldOfStudy(McGill University, d) $\wedge$ $\neg$Language(Nico, d) $\wedge$ FieldOfStudy(e, d) & Who studies the field that is studied by McGill University, but is not spoken by Nico? \\ \midrule
pin & \includegraphics[width=2cm]{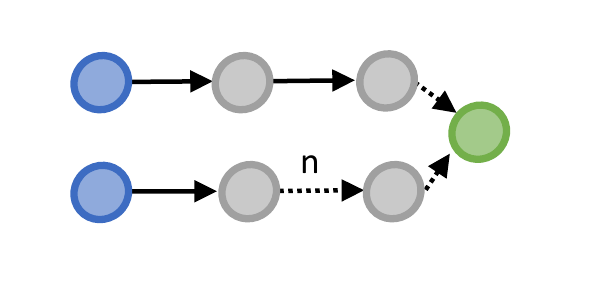} & q =?e : $\exists$ a : Rel\_1(A, a) $\wedge$ Rel\_2(a, e) $\wedge$ $\neg$Rel\_3(B, e) & q =?e : $\exists$ a : Country(a, USA) $\wedge$ Student(a, e) $\wedge$ $\neg$Film(e, Malcolm X) & Who was a student of a university in United States, but did not film Malcolm X? \\ \midrule
pni & \includegraphics[width=2cm]{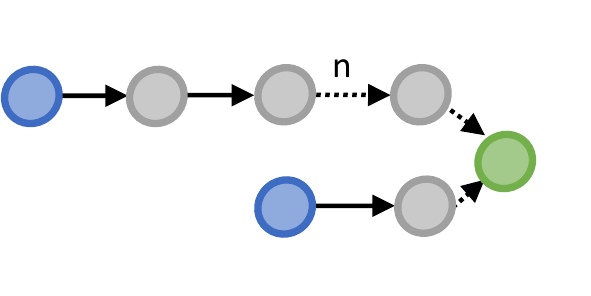} & q =?e : $\exists$ a : Rel\_1(A, a) $\wedge$ $\neg$Rel\_2(a, e) $\wedge$ Rel\_3(B, e) & q =?e : $\exists$ a : SymptomOf(dyspnea, a) $\wedge$ $\neg$CauseOfDeath(e, a) $\wedge$ Religion(e, Catholicism) & Who believed in Catholicism and did not die from the disease that has the symptom of dyspnea? \\ 
\bottomrule
\end{tabular}
\end{table*}


\end{document}